\documentclass[conference]{IEEEtran}
\IEEEoverridecommandlockouts

\usepackage{cite}
\usepackage{amsmath,amssymb,amsfonts}
\usepackage{algorithmic}
\usepackage{graphicx}
\graphicspath{{../}}
\usepackage{textcomp}
\usepackage{xcolor}
\def\BibTeX{{\rm B\kern-.05em{\sc i\kern-.025em b}\kern-.08em
    T\kern-.1667em\lower.7ex\hbox{E}\kern-.125emX}}
    
\usepackage{bm}
\usepackage{booktabs}
\usepackage[caption=false,font=footnotesize]{subfig}
\usepackage{url}

\begin{document}

\title{%
  {\normalfont\small\itshape
    Accepted at the 2026 IEEE International Conference on Data Mining (ICDM 2026)%
  }\\[0.6em]
  BiGSyn: A Bi-Level Heterogeneous Graph Contrastive Learning Framework
  for Interpretable Drug Synergy Prediction%
  \thanks{\textsuperscript{$\ast$} Corresponding author.}%
  \thanks{\textcopyright{} 2026 IEEE. Personal use of this material is
  permitted. Permission from IEEE must be obtained for all other uses,
  in any current or future media, including reprinting/republishing this
  material for advertising or promotional purposes, creating new collective
  works, for resale or redistribution to servers or lists, or reuse of any
  copyrighted component of this work in other works.}%
}


\author{
	\IEEEauthorblockN{
		Senquan Wang\textsuperscript{1}, 
		Hui Fang\textsuperscript{2$\ast$}, 
        Haoran Nan\textsuperscript{1}, 
        Xingwei Wang\textsuperscript{3}, 
		Qiang He\textsuperscript{3$\ast$}} 
	\IEEEauthorblockA{\textsuperscript{1}College of Medicine and Biological Information Engineering, Northeastern University, Shenyang, China}
	\IEEEauthorblockA{\textsuperscript{2}School of Computing and Artificial Intelligence, Shanghai University of Finance and Economics, Shanghai, China} 
	\IEEEauthorblockA{\textsuperscript{3}College of Computer Science and Engineering, Northeastern University, Shenyang, China}
    \IEEEauthorblockA{senquanwang@foxmail.com, fang.hui@mail.shufe.edu.cn, \{nhr0211, net2020zjq\}@163.com, heqiang@bmie.neu.edu.cn}
} 

\maketitle

\begin{abstract}
    Drug combination therapy has become a mainstream of modern medicine, with the potential to enhance efficacy, reduce toxicity, and overcome drug resistance. 
    However, identifying synergistic drug combinations remains challenging, as it requires modeling complex interactions across multiple biological scales---from atom-level molecular structures to entity-level relations in biological networks. 
    Existing approaches typically model these levels in isolation, overlooking complementary signals and long-range dependencies. 
    To address these limitations, we propose \textbf{BiGSyn}, a \textbf{Bi}-level heterogeneous \textbf{G}raph contrastive learning framework for drug \textbf{Syn}ergy prediction. 
    Specifically, BiGSyn leverages heterogeneous graph transformers to extract high-order dependencies from molecular graphs and biological networks, and introduces a bi-level contrastive learning objective to bridge the cross-level semantic gap. 
    Extensive experiments on two benchmark datasets show that BiGSyn consistently outperforms state-of-the-art baselines. 
    Moreover, its attention mechanism highlights key molecular substructures and biological interactions driving synergy, providing interpretable insights. 
    
\end{abstract}

\begin{IEEEkeywords}
Drug Synergy Prediction, Heterogeneous Graph Learning, Contrastive Learning
\end{IEEEkeywords}

\section{Introduction}

Drug combination therapy---the concurrent use of two or more therapeutic agents---plays a crucial role in modern medicine due to its advantages over monotherapy, such as enhanced efficacy, reduced side effects and toxicity, and the ability to overcome drug resistance~\cite{lehar2009synergistic,jaaks2022effective,jin2023rational}. Depending on whether the combined effect exceeds the sum of individual effects, drug combinations are classified as synergistic, additive, or antagonistic. 
In recent years, researchers have paid increasing attention to drug synergy prediction, i.e., identifying synergistic combinations that exhibit enhanced therapeutic efficacy, as it directly enables regimen optimization and accelerates the discovery of effective treatments~\cite{abbasi2024new,pourmousa2025ai}. 
 
\begin{figure}[t]
    \centering
    \includegraphics[width=1\linewidth]{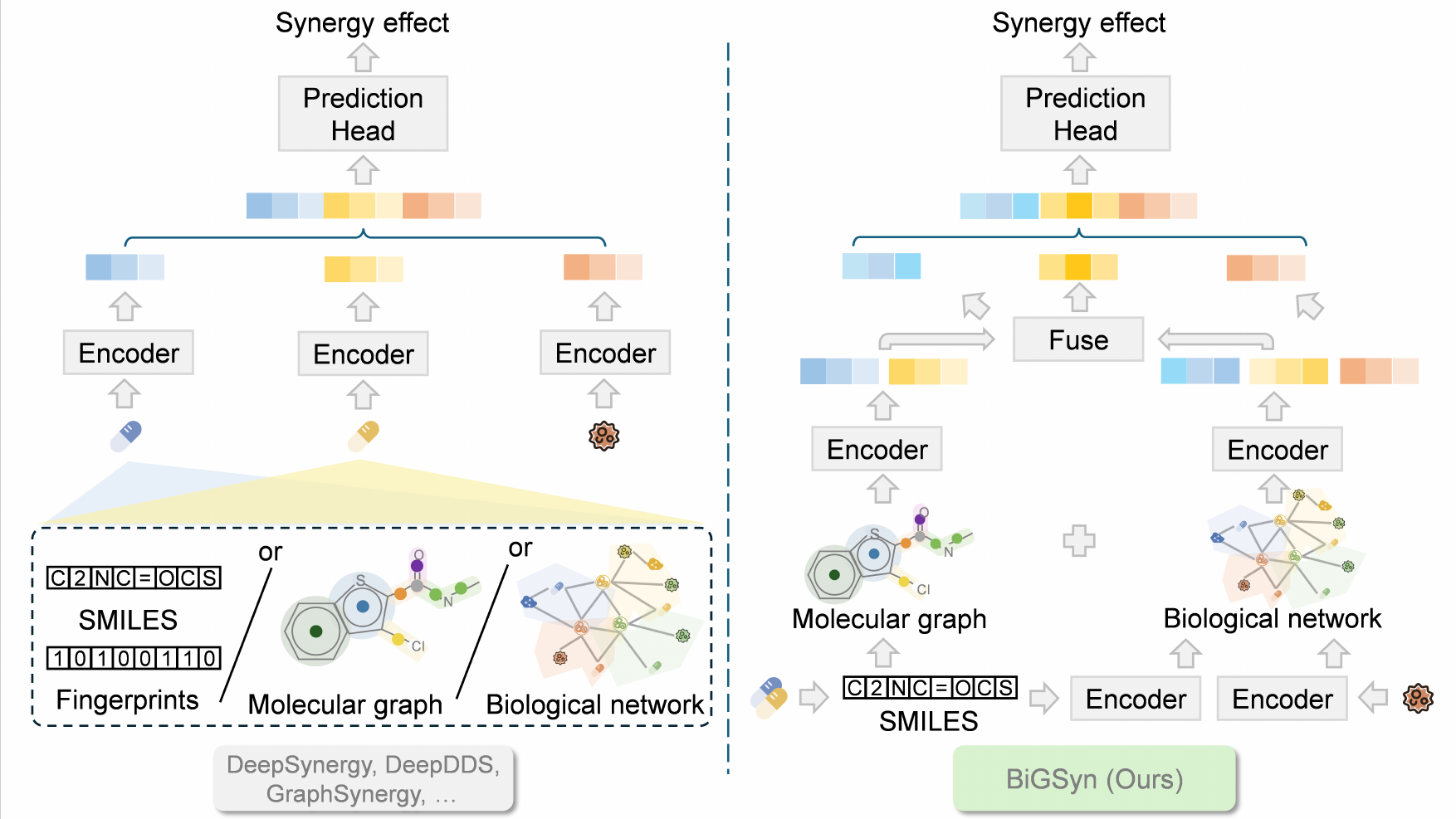}
    \caption{BiGSyn vs. existing models for drug synergy prediction.}
    \label{fig_comparison}
\end{figure}

However, identifying synergistic combinations experimentally is impractical due to the combinatorial explosion of potential pairs and the limited understanding of underlying mechanisms~\cite{sun2013high,han2017synergistic,wang2022deepdds}. Consequently, data-driven computational approaches have emerged as promising alternatives. Early methods relied on sequential or vector representations, encoding chemical structures using Simplified Molecular Input Line Entry System (SMILES) strings or molecular fingerprints~\cite{preuer2018deepsynergy,kuru2021matchmaker,xu2023dffndds}. While compact, these representations fail to capture the topological and structural details critical to pharmacological behavior. With the rise of graph neural networks (GNNs), molecular graphs have become a prominent representation for modeling drug structures. 
By representing atoms as nodes and chemical bonds as edges, molecular graphs explicitly capture atom-level topology and bond relationships~\cite{zhang2022predictingPLB,besharatifard2024review}. 
Since molecular structure largely determines physicochemical properties and pharmacological actions~\cite{mao2016chemical}, molecular-graph-based models have been widely applied to molecular property~\cite{zhao2024molecular} and drug response prediction~\cite{nguyen2021graph}. Parallel to structural modeling, contextual representations derived from biological networks model interactions among drugs, targets, and cellular context, which strongly influence efficacy and toxicity~\cite{cheng2019network,yang2021graphsynergy,Li2023discovering}. 

Despite these advances, most existing approaches operate on a single level---either molecular graphs or biological networks---ignoring complementary hierarchical signals, as shown in Fig.~\ref{fig_comparison}. We identify two main challenges: (1) \textbf{Single-level modeling}. Methods limited to one graph type cannot fully capture interactions spanning chemical and biological spaces. Specifically, methods focusing solely on molecular graphs miss the broader biological context in which drugs act, while those using only biological networks overlook fine-grained structural details. (2) \textbf{Cross-level semantic gap}. Molecular and biological graphs differ not only in scale but also in heterogeneity, as they contain different types of nodes and edges~\cite{liu2023recent}. This structural discrepancy complicates joint representation learning, making effective fusion difficult and often leading to suboptimal predictions.

To tackle these challenges, we propose BiGSyn, a bi-level heterogeneous graph contrastive learning framework for drug synergy prediction. At the micro-level, each drug is represented as a fine-grained graph of atoms and bonds. At the macro-level, a heterogeneous biological network integrates drugs, cell lines, proteins, and tissues, along with multiple types of interactions. We employ heterogeneous graph transformers to extract fine-grained atom-level dependencies from molecular graphs and high-order entity-level dependencies from the biological network. To integrate these two levels, we design a bi-level contrastive learning objective that aligns molecular and biological representations, bridging the semantic gap and producing more expressive embeddings. Finally, the integrated drug embeddings are combined with cell line features and fed into a learnable classifier to predict synergistic drug combinations.

The contributions of this work are summarized as follows:
\begin{itemize}
    \item We propose a novel framework for drug synergy prediction that jointly models molecular and biological graphs as heterogeneous graphs, capturing complementary signals across scales.
    \item We leverage heterogeneous graph transformers to capture micro-level chemical interactions and macro-level biological dependencies, while multi-head attention identifies key substructures and critical interactions.
    \item We introduce a bi-level contrastive learning objective to align representations across hierarchies, yielding more expressive drug embeddings.
    \item Extensive experiments demonstrate superior performance and generalization under drug-pair- and cell-line-level label partitioning. A case study further validates the model's ability to infer potential synergistic drug combinations.
\end{itemize}

\section{Related Work}

\subsection{Drug Synergy Prediction}

With the rapid accumulation of multi-omics data and the availability of large curated datasets, data-driven deep learning approaches have become increasingly dominant in drug synergy prediction. 
Early approaches relied on sequence-based representations, where molecular fingerprints or SMILES strings encode chemical structures as fixed-length vectors or character sequences. 
These sequential representations offer compact encodings of molecular composition and have been widely adopted in deep learning models such as DeepSynergy~\cite{preuer2018deepsynergy} and SMILESynergy~\cite{zhang2023smilesynergy}, which employ deep neural networks to learn drug representations from fingerprints or SMILES-derived features. 
Recent work has also explored pre-trained language models for encoding SMILES strings and protein sequences~\cite{wei2022deeplpi,xu2023dffndds}, demonstrating the value of sequence-derived semantic features. 
However, such sequence-based encodings cannot explicitly capture atom-level chemical relationships and overlook the complex relational dependencies among drugs and biological entities. 

To address this, graph-based approaches have been developed, as they can naturally capture and explicitly model such intricate interactions~\cite{wang2022deepdds,zhao2024molecular,zeng2025graphfusion}. 
Structural representations via molecular graphs have become prominent, capturing atom-level chemical topology and bond relationships more explicitly. Nevertheless, these approaches focus on drug-centric representations and miss the biological context in which drugs operate. 
More recently, contextual representations from biological knowledge networks have gained attention, modeling the interplay among drugs, their molecular targets and the cellular context to capture system-level interactions~\cite{zhang2023kgansynergy,cheng2024hansynergy}. However, such network-based methods lack fine-grained structural information about chemical properties.


\subsection{Graph-based Representation Learning}

Graph neural networks (GNNs) have shown exceptional capability in learning expressive representations from structured and relational data. 
For molecular graphs, GNN variants such as GCN, GAT and GIN have been widely applied to model and learn drug embeddings~\cite{wang2022deepdds,lu2025context,zhang2025mosaic}. 
These models capture local topological patterns and chemical semantics, yet they typically rely on homogeneous graphs, limiting their ability to model multi-entity biological systems. 
In contrast, heterogeneous graph neural networks (HGNNs) extend this paradigm to handle multiple node and edge types, enabling the modeling of complex biomedical interactions~\cite{zhang2023kgansynergy,cheng2024hansynergy,Wang2025Fusing}. 
Recently, graph transformers have further advanced this field by incorporating the attention mechanisms of Transformers to capture long-range dependencies, showing strong capabilities in drug combination prediction~\cite{alam2024unlocking,sun2025enhancing,liu2025mgtnsyn}. 
Nevertheless, most existing graph-based methods focus on a single level of interactions---either molecular graphs or biological networks---and remain constrained to local neighborhood aggregation. 
Moreover, the structural discrepancy and semantic gap between atom-level molecular graphs and entity-level biological networks pose a challenge for effective joint representation learning. 
Such limitations hinder their ability to fully capture the intricate cross-level relationships between drug molecular structures and biological knowledge networks, highlighting the need for a unified framework that can jointly model and align representations across different scales.

\section{Methodology}

\begin{figure*}[t]
        \centering\includegraphics[width=0.8\textwidth]{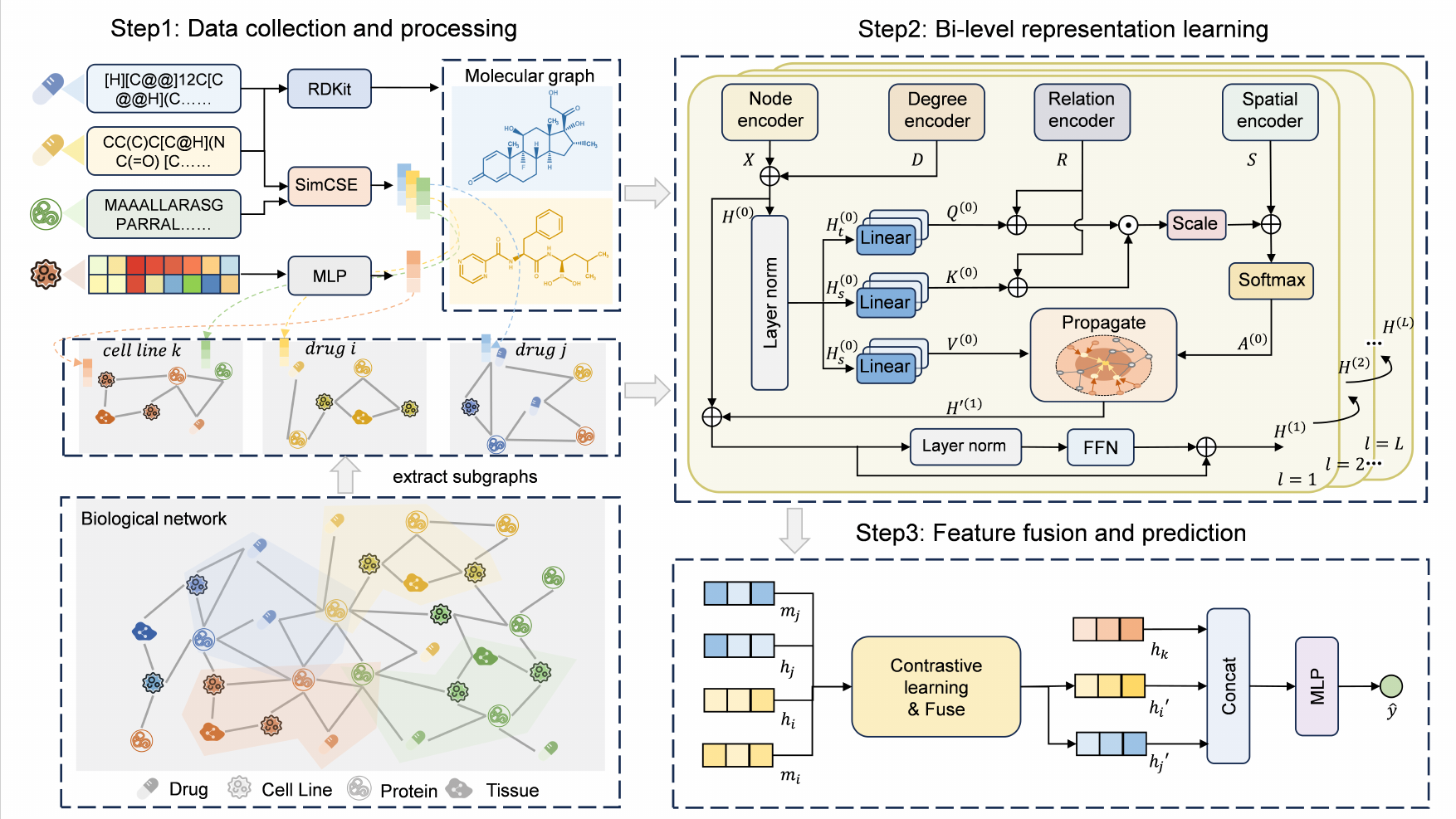}
        \caption{The overall framework of BiGSyn.}
        \label{fig:architecture}
\end{figure*}

\subsection{Preliminaries}

Given a drug set $D=\{d_1,d_2,...,d_n\}$ and a cell line set $C=\{c_1,c_2,...,c_m\}$, a drug combination sample is denoted as $(d_i,d_j,c_k,y)$, where $(d_i,d_j)$ represents a drug pair, $c_k$ denotes a specific cell line, and a binary label $y\in\{0,1\}$ indicates whether the drug combination exhibits a synergistic effect under the given cellular context. The goal of our work is to leverage the training data to learn a mapping function $f:(d_i,d_j,c_k)\rightarrow[0,1]$ that predicts the probability that drugs $d_i$ and $d_j$ are synergistic when applied to cell line $c_k$. 

\subsection{Overview}

The overall architecture of BiGSyn is illustrated in Fig.~\ref{fig:architecture}. 
Specifically, we first construct a macro-level heterogeneous biological network that integrates drugs, cell lines, target proteins and tissues with multiple types of relations. 
To mitigate computational overhead and reduce potential noise from the full graph, we extract a set of subgraphs centered on drug and cell-line nodes via random walks. 
Meanwhile, sequence-based encoders are employed to initialize node features in the biological network using modality-specific information, while molecular graphs are constructed from drug SMILES strings to characterize atom-level interactions. 
Subsequently, heterogeneous graph transformers are employed to learn representations from both the macro-level biological subgraphs and the micro-level molecular graphs. 
Afterwards, a bi-level contrastive learning objective is introduced to align and fuse the drug embeddings from these two levels. 
Finally, the integrated drug representations are concatenated with cell line embeddings and fed into a learnable classifier for synergy prediction. 
The entire framework is trained end-to-end, jointly optimizing contrastive and predictive objectives to achieve robust and interpretable drug synergy prediction. 

\subsection{Biological Graph Construction}

To capture structural and contextual information, we construct complementary heterogeneous biological graphs with sequence-based feature initialization. 

\subsubsection{Molecular Graph Construction}

For each drug $d$, we employ RDKit~\cite{landrum2013rdkit} to convert its SMILES representation into a graph $\mathcal{G}_M=(\mathcal{A},\mathcal{B},\mathcal{T}_a,\mathcal{T}_b)$, where $\mathcal{A}$ and $\mathcal{B}$ denote the sets of atoms and chemical bonds, respectively, and $\mathcal{T}_a$ and $\mathcal{T}_b$ represent their corresponding types. 
The initial atom features encode physicochemical properties, including atom type, valence, and aromaticity. Each chemical bond is treated as a typed edge, where the bond type is encoded as an edge relation. 

\subsubsection{Knowledge Network Construction}

As proteins play pivotal roles in signaling, metabolism, immunity, and gene regulation networks~\cite{zhang2023kgansynergy,Li2023discovering}, we follow the setting of~\cite{yang2021graphsynergy} to construct a biological interaction knowledge graph based on the protein-protein interaction (PPI) network. 
In this graph, nodes correspond to drugs, target proteins, cell lines, and tissues, while edges represent their biological or functional associations. 
The network integrates multiple types of interactions, including protein-protein, drug-protein, cell line-protein, and cell line-tissue relationships. 

\subsubsection{Sequential Feature Initialization}

To enrich the biological semantics of nodes in the knowledge graph, we initialize node representations with sequence-level information. Following~\cite{cheng2024hansynergy}, we employ its pre-trained SimCSE-based encoder to generate fixed-length embeddings for SMILES strings and amino acid sequences, which are used as the initial features of drug and protein nodes, respectively. These sequence-derived features provide complementary semantic information to the molecular structural representations and graph-contextual information learned by our model. For cell line nodes, we initialize features using gene expression profiles from the Cancer Cell Line Encyclopedia (CCLE)~\cite{barretina2012cancer}, while tissue nodes are initialized with one-hot encodings. 

Consequently, we obtain a heterogeneous biological interaction knowledge network $\mathcal{G}_B=(\mathcal{V},\mathcal{E},\mathcal{T}_v,\mathcal{T}_e)$, where $\mathcal{V}$ and $\mathcal{E}$ denote the sets of nodes and edges, and $\mathcal{T}_v$ and $\mathcal{T}_e$ indicate their types. 
However, the large scale of this graph imposes substantial computational and memory burdens if processed in its entirety. 
Moreover, the Transformer architecture is constrained by long input sequences, and excessive noise in large graphs may further degrade model performance. 
To mitigate these issues, we adopt a random-walk--based subgraph sampling strategy~\cite{grover2016node2vec} to extract a set of subgraphs centered on drug and cell-line nodes. 

Finally, to enhance spatial dependency modeling, we augment these biological graphs by computing shortest-path distances between node pairs and incorporating them as additional edge relations. 

\subsection{Bi-level Representation Learning}

To capture fine-grained atom-level features from molecular graphs and coarse-grained entity-level features from biological subgraphs, we employ two heterogeneous graph transformers. 
Each transformer is equipped with multi-aspect graph encoders and attention-based message propagation layers to model complex dependencies within graphs. 

\subsubsection{Multi-Aspect Graph Encoders}

Inspired by~\cite{ying2021transformers} and~\cite{su2024dual}, we incorporate multi-aspect graph encoders into the heterogeneous graph transformer to better accommodate the non-Euclidean characteristics of graph-structured data, as illustrated in Fig.~\ref{fig:architecture}. 
The node encoder projects features of different node types into a unified latent space. 
Formally, for a heterogeneous graph $\mathcal{G}_*=(\mathcal{V},\mathcal{E},\mathcal{T}_v,\mathcal{T}_e)$, where $\mathcal{V}$ is the set of nodes with a node type mapping function $\Phi:\mathcal{V}\rightarrow \mathcal{T}_v$, and $\mathcal{E}$ is the set of edges with an edge type mapping function $\Psi:\mathcal{E}\rightarrow \mathcal{T}_e$, the projection is defined as: 
\begin{equation}
\bm{x}_v=\bm{W}_{\Phi(v)}\bm{x}_v^0,
\end{equation}
where $\bm{W}_{\Phi(v)}$ is a type-specific projection matrix shared by all nodes of type $\Phi(v)$, $\bm{x}_v^0$ and $\bm{x}_v$ are the original and projected features of node $v$, respectively. 
The projected features of all nodes form $\bm{X}\in\mathbb{R}^{|\mathcal{V}|\times d}$, with $d$ denoting the hidden dimension. 
The degree encoder incorporates centrality information by embedding node degrees: 
\begin{equation}
\bm{d}_v=\phi(deg(v)),
\end{equation}
where $deg(v)$ denotes the degree of node $v$, and $\phi(\cdot)$ is an embedding function. The degree embeddings of all nodes are represented as $\bm{D}\in\mathbb{R}^{|\mathcal{V}|\times d}$. 
The relation encoder models semantic differences between edge types by assigning a unique embedding to each edge: 
\begin{equation}
\bm{r}_e=\psi(\Psi(e)),
\end{equation}
where $\Psi(e)$ maps edge $e$ to its type, $\psi(\cdot)$ is an embedding layer, and the embeddings of all edges are collectively represented as $\bm{R}\in\mathbb{R}^{|\mathcal{E}|\times d}$. 
The spatial encoder captures structural relations by encoding the shortest path distance between two nodes: 
\begin{equation}
s_e=\mathrm{ReLU}(\bm{W}_2\mathrm{ReLU}(\bm{W}_1 \zeta(e))),
\end{equation}
where $\zeta(e)$ denotes the shortest path distance of edge $e$, $\bm{W}_1\in\mathbb{R}^{1\times d}$ and $\bm{W}_2\in\mathbb{R}^{d\times 1}$ are learnable weights. The spatial relations are collectively represented as $\bm{S}\in\mathbb{R}^{|\mathcal{E}|\times 1}$. 

\subsubsection{Weighted Message Propagation}

After obtaining node-, degree-, relation-, and distance-aware representations $\bm{X}$, $\bm{D}$, $\bm{R}$, and $\bm{S}$, we employ a multi-head attention mechanism to perform weighted message propagation over heterogeneous edges. 
Specifically, at the $l$-th layer, the hidden representations of nodes $\bm{H}^{(l)} \in \mathbb{R}^{|\mathcal{V}|\times d}$ are projected into $\bm{Q}$, $\bm{K}$, $\bm{V}$ for each attention head via learnable weight matrices $\bm{W}_q\in\mathbb{R}^{d\times d_k}$, $\bm{W}_k\in\mathbb{R}^{d\times d_k}$ and $\bm{W}_v\in\mathbb{R}^{d\times d_v}$ respectively: 
\begin{gather}
\bm{Q}^{(l)(i)}=\bm{H}^{(l)}\bm{W}_q^{(l)(i)}, \ \bm{K}^{(l)(i)}=\bm{H}^{(l)}\bm{W}_k^{(l)(i)},\\
\nonumber \bm{V}^{(l)(i)}=\bm{H}^{(l)}\bm{W}_v^{(l)(i)}, \ l=0,1,\dots,L-1, \ i=1,2,\dots,N_h,
\end{gather}
where $L$ denotes the total number of layers, and $d_k=d_v=d/N_h$ with $N_h$ being the number of attention heads. The initial node representations are $\bm{H}^{(0)}=\bm{X}+\bm{D}$. 
To propagate messages along heterogeneous edges, we construct edge-aligned representations by gathering node features according to source–target pairs. 
For each edge $(s,t)\in\mathcal{E}$, query is obtained from the target node, while key and value are obtained from the source node. 
Accordingly, $\bm{Q}_t^{(l)(i)}$, $\bm{K}_s^{(l)(i)}$, $\bm{V}_s^{(l)(i)}$ are all of shape $\mathbb{R}^{|\mathcal{E}|\times d}$. 

To enhance the model’s understanding of the semantic relationships of edges and the structural information of the graph, we incorporate the relation embeddings $\bm{R}$ and spatial encodings $S$ in the computation of attention weights: 
\begin{gather}
\bm{A}^{(l)(i)} =\mathrm{softmax}\left(\frac{\bm{Q}_t^{(l)(i)\prime}\bm{K}_s^{(l)(i)\prime\top}}{\sqrt{d_k}} + \bm{S}\right), \\
\nonumber \bm{Q}_t^{(l)(i)\prime}=(\bm{Q}_t^{(l)(i)}+\bm{R}^{(i)}), \ \bm{K}_s^{(l)(i)\prime}=(\bm{K}_s^{(l)(i)}+\bm{R}^{(i)}).
\end{gather}
The computed attention weights $\bm{A}^{(l)(i)}$ are used to perform message passing and update the node representations, where the feature of the target node $t$ is updated as: 
\begin{equation}
\bm{h}_{t}^{(l+1)(i)}=\bm{h}_{t}^{(l)(i)}+\sum_{s\in\mathcal{N}(t)}\alpha_{st}^{(l)(i)}\bm{v}_s^{(l)(i)}, \ i=1,2,...,N_h,
\end{equation}
where $\mathcal{N}(t)$ denotes the set of neighbors of node $t$, and $\alpha_{st}^{(l)(i)}$ is the attention weight of the $i$-th head corresponding to the edge $(s,t)$. 
The final node representation is obtained by concatenating the outputs from all heads: 
\begin{equation}
\bm{h}_t^{(l+1)}=\Bigg\Vert_{i=1}^{N_h} \bm{h}_t^{(l+1)(i)},
\end{equation}
and the hidden states of all nodes are collectively represented as $\bm{H}^{(l+1)}$. 

The whole update process can be formally characterized as follows: 
\begin{align}
\bm{H}^{\prime(l+1)} & =\mathrm{MHA}(\mathrm{LN}(\bm{H}^{(l)}))+\bm{H}^{(l)}, \\
\bm{H}^{(l+1)} & =\mathrm{FFN}(\mathrm{LN}(\bm{H}^{\prime(l+1)}))+\bm{H}^{\prime(l+1)},
\end{align}
where $\mathrm{MHA}$ and $\mathrm{FFN}$ denote the multi-head attention and feed-forward network modules, respectively, and $\mathrm{LN}$ represents layer normalization, 
following the Pre-LN Transformer variant~\cite{xiong2020layer}. 

\subsubsection{Graph-level and Node-level Features Extraction}

The molecular graph-level representation of drug $d_i$ is obtained via global mean pooling over all atom nodes: 
\begin{equation}
\bm{m}_i=\mathrm{MeanPool}(\{\bm{h}_a^{(L)}|a\in \mathcal{A}\}),
\end{equation}
where $\mathcal{A}$ denotes the set of atom nodes in the molecular graph of $d_i$. 

For the biological subgraphs, the node-level representations of drugs or cell lines are directly extracted from the updated node feature matrix according to their corresponding indices: 
\begin{equation}
\bm{h}_i=\bm{H}^{L}[idx],
\end{equation}
where $idx$ denotes the index of entity $i$ within its biological subgraph. 

\subsection{Contrastive Learning and Fusion}

For each drug, we obtain two representations from the molecular graph and biological subgraph via the heterogeneous graph transformer. To align them and stabilize training, we adopt a contrastive learning framework
~\cite{chen2025knowledge}, where embeddings of the same drug across the two levels serve as positive pairs, and embeddings of different drugs form negative pairs. 
For a mini-batch of size $N$, the contrastive loss for drug $d_i$ is formally defined as: 
\begin{align}
    \nonumber
    \mathcal{L}_i&=-\log\frac{e^{<\bm{m}_i,\bm{h}_i>/\tau}}{\sum_{j=1,j\neq i}^N e^{<\bm{m}_i,\bm{m}_j>/\tau}+\sum_{j=1}^N e^{<\bm{m}_i,\bm{h}_j>/\tau}}\\
    &-\log\frac{e^{<\bm{h}_i,\bm{m}_i>/\tau}}{\sum_{j=1,j\neq i}^N e^{<\bm{h}_i,\bm{h}_j>/\tau}+\sum_{j=1}^N e^{<\bm{h}_i,\bm{m}_j>/\tau}},
\end{align}
where $\bm{m}_i$ and $\bm{h}_i$ denote the representations of drug $d_i$ obtained from the molecular graph and biological subgraph, respectively. $<\cdot,\cdot>$ denotes the inner product to measure the similarity between two representations, and $\tau$ is the temperature hyperparameter which controls the trade-off between emphasizing hard negatives for sharper cross-level alignment and maintaining stable optimization in the presence of semantically similar drugs. 
The overall contrastive learning objective is then averaged across all drugs in the batch: 
\begin{equation}
    \mathcal{L}_{CL}=\frac{1}{N}\sum_{i=1}^N\mathcal{L}_i.
\end{equation}

To further distill the essential information of drugs from the two levels, we concatenate their representations and feed them into an MLP to obtain the fused representation $\bm{h}'_i\in \mathbb{R}^d$ of drug $d_i$: 
\begin{equation}
    \bm{h}'_i=\mathrm{MLP}\left([\bm{m}_i\parallel \bm{h}_i]\right),
\end{equation}
where $\bm{m}_i$ and $\bm{h}_i$ denote the molecular graph and biological subgraph representations of $d_i$, respectively, and $[\cdot \parallel \cdot]$ represents the concatenation operation. 

\subsection{Prediction and Training}

Based on the previously obtained cell line embeddings and the fused drug representations, we perform the final drug synergy prediction, which can be formulated as a binary classification task. For any drug–drug–cell line triplet $(d_i,d_j,c_k)$, we concatenate their embeddings and feed them into an MLP to estimate the probability of synergy: 
\begin{equation}
    \hat{y}_{(i,j,k)}=\mathrm{MLP}([\bm{h}'_i\parallel \bm{h}'_j\parallel \bm{h}_k]),
\end{equation}
where $\bm{h}'_i$ and $\bm{h}'_j$ denote the fused representations of drugs $d_i$ and $d_j$, $\bm{h}_k$ is the representation of cell line $c_k$. 
The cross-entropy loss is adopted as the objective function: 
\begin{equation}
    \mathcal{L}_{CE}=-\frac{1}{N}\sum_{(i,j,k)}\left[y\log\hat{y}+(1-y)\log(1-\hat{y})\right],
\end{equation}
where $y\in\{0,1\}$ denotes the ground truth label. 
To alleviate overfitting, we further introduce an L2 regularization term. 
The final loss function is thus defined as: 
\begin{equation}
    \mathcal{L}=\mathcal{L}_{CE}+\delta\mathcal{L}_{CL}+\lambda\|\Theta\|_2^2,
\end{equation}
where $\Theta$ denotes the set of model parameters, $\delta$ is a trade-off hyperparameter for the contrastive loss, and $\lambda$ is the coefficient for L2 regularization. 

\section{Experiments}

\subsection{Experimental Setup}

\subsubsection{Datasets}

\begin{table}[t]
\caption{Statistics of the datasets after processing.\label{tab:dataset}}
\begin{center}
\begin{tabular*}{\columnwidth}{@{\extracolsep\fill}lrr@{\extracolsep\fill}}
\toprule
\textbf{} & \textbf{Oncology-Screen} & \textbf{DrugCombDB} \\
\midrule
Drugs        & 21    & 572   \\
Cell lines   & 29    & 65    \\
Tissues      & 6     & 13    \\
Combinations & 1,960  & 30,630 \\
Positive     & 1,044  & 15,318 \\
Negative     & 916   & 15,312 \\
\bottomrule
\end{tabular*}
\end{center}
\end{table}

We extensively evaluate BiGSyn on two benchmark datasets of different scales, namely Oncology-Screen~\cite{o2016unbiased} and DrugCombDB~\cite{liu2020drugcombdb}.
\textbf{Oncology-Screen} is a relatively small dataset containing 22,737 combinations across 583 drugs, 39 cancer cell lines, and 6 tissue types. 
\textbf{DrugCombDB} is a large-scale dataset that integrates drug combination data from multiple sources, comprising 448,555 combinations covering 2,887 drugs, 124 cancer cell lines, and 14 tissue types. 
Following the preprocessing protocol in~\cite{liu2020drugcombdb}, we label the top 25\% of drug combinations ranked by synergy scores as synergistic and the bottom 25\% as antagonistic. 
The statistics of the processed datasets are summarized in Table~\ref{tab:dataset}. 
We construct a heterogeneous biological knowledge graph for each dataset. 
The detailed statistics of these networks are summarized in Table~\ref{tab:network}. 

\begin{table}[t]
\caption{Statistics of the heterogeneous biological networks.\label{tab:network}}%
\begin{center}
\begin{tabular*}{\columnwidth}{@{\extracolsep\fill}lrr@{\extracolsep\fill}}
\toprule
\textbf{}        & \textbf{Oncology-Screen} & \textbf{DrugCombDB}\\
\midrule
\#Nodes          & 15,936                   & 16,530             \\
\#Links          & 452,354                  & 488,496            \\
\#Relations      & 8                        & 8                  \\
\bottomrule
\end{tabular*}
\end{center}
\end{table}

\subsubsection{Baselines}

To validate the superiority of our proposed approach, we compared it with six representative deep learning methods. 
DeepSynergy~\cite{preuer2018deepsynergy} is a non-graph-based method that employs a fully connected deep neural network. 
GraphSynergy~\cite{yang2021graphsynergy}, KGANSynergy~\cite{zhang2023kgansynergy} and DeepTraSynergy~\cite{rafiei2023deeptrasynergy} are graph-based methods that leverage biological interaction networks for drug synergy prediction. 
SynergyGTN~\cite{alam2024unlocking} and GraphTranSynergy~\cite{sun2025enhancing} focus on molecular graphs and utilize graph transformer architectures to learn drug representations. 

\subsubsection{Evaluation Protocols}

To comprehensively evaluate the generalization ability of our model, we employed three types of 5-fold cross-validation (CV) strategies. 
Specifically, 90\% of the samples from each dataset were randomly selected as the CV set, while the remaining 10\% were held out as an independent test set. 
Within the CV set, we designed three scenarios: 
(1) sample-level CV (\textbf{CV1}), where drug–drug–cell line triplets were randomly partitioned; 
(2) drug pair-level CV (\textbf{CV2}), where folds were split by drug pairs; and 
(3) cell line-level CV (\textbf{CV3}), where folds were split by cell lines. 
This setup allows us to assess model performance under different generalization scenarios, from the basic random split to stricter cases where the model must predict synergy labels associated with cell lines or drug pairs not included in training.

\subsubsection{Implementation Details}

At both the molecular and biological network levels, we employed two-layer heterogeneous graph transformers with embedding dimension of $d=256$, dropout $\rho=0.2$, and four attention heads. The sampled subgraph size was set to 32. Drug feature fusion used a three-layer MLP with hidden sizes of $[4d, 2d, d]$. For contrastive learning, we set the temperature parameter $\tau=0.1$ and the loss weight $\delta=0.1$. The prediction head was implemented as a three-layer MLP with hidden sizes of $[4d, 8d, 2]$ and a dropout rate of 0.2. The learning rates were 0.001 for the Oncology-Screen dataset and 0.0001 for the DrugCombDB dataset, with 200 and 400 training epochs, respectively. We adopted the Adam optimizer with a weight decay of 0.0001, batch size of 128, and an early stopping strategy with a patience of 20 epochs. All experiments were implemented in PyTorch and conducted on a server equipped with an NVIDIA Tesla V100 GPU. 
The hyper-parameter sensitivity analysis can be found in Section~\ref{sec:param}. 
Our code and data are available at: \url{https://github.com/senquanwang/BiGSyn}.

\subsection{Comparison with Existing Methods}

\begin{table*}[t]
\caption{Independent test performance (in \%) comparison across datasets and models on sample-level CV. The best results are indicated in \textbf{bold} while the sub-optimal results are \underline{underlined}. Statistical significance from paired $t$-test is denoted as \text{***}, \text{**}, and \text{*} for $p < 0.01$, $p < 0.05$, and $p < 0.1$, respectively.\label{tab:CV1}}
\begin{center}
\tabcolsep=0pt
\begin{tabular*}{\textwidth}{@{\extracolsep{\fill}}lrrrrrrrrrrrr@{\extracolsep{\fill}}}  
\toprule%
& \multicolumn{6}{@{}c@{}}{Oncology-Screen} & \multicolumn{6}{@{}c@{}}{DrugCombDB} \\
\cmidrule(lr){2-7}\cmidrule(lr){8-13}
Method          & ACC    & Precision & Recall & AUC    & AUPR   & F1        & ACC    & Precision & Recall & AUC    & AUPR   & F1 \\
\midrule
DeepSynergy     & 77.15 & 84.37    & 70.28 & 85.81 & 83.42 & 76.46    & 81.39 & 83.48    & 78.31 & 89.7  & 89.43 & 80.79\\
DeepTraSynergy  & 76.44 & 77.70    & 78.28 & 76.31 & 72.41 & 77.97    & 74.64 & 75.86    & 72.33 & 74.64 & 68.69 & 74.02\\
GraphSynergy    & \underline{84.36} & \underline{86.32} & 84.00 & \underline{91.72} & 92.41 & \underline{85.11} & \underline{83.90} & \underline{83.60} & \underline{84.34} & \underline{91.65} & \underline{91.69} & \underline{83.97}\\
KGANSynergy     & 83.55 & 85.86    & 82.85 & 89.24 & 88.95 & 84.28    & 79.24 & 78.30     & 80.92 & 86.93 & 86.05 & 79.57\\
SynergyGTN      & 83.45 & 83.55    & \underline{85.90}  & 90.12 & 91.54 & 84.67    & 79.79 & 82.06    & 76.38 & 88.55 & 88.73 & 79.07\\
GraphTranSynergy& 82.84 & 83.86 & 84.00 & 91.25 & \underline{93.03} & 83.90 & 65.73 & 64.84 & 67.87 & 71.19 & 70.13 & 66.18\\
BiGSyn(Ours)& \textbf{88.83} & \textbf{91.02} & \textbf{87.80} & \textbf{94.37} & \textbf{95.18} & \textbf{89.34} & \textbf{86.76} & \textbf{86.94} & \textbf{86.54} & \textbf{93.83} & \textbf{93.83} & \textbf{86.73}\\
\midrule
Improve.       & 4.47 & 4.70 & 1.90 & 2.65 & 2.15 & 4.23 & 2.86 & 3.34 & 2.21 & 2.18 & 2.14 & 2.76\\
Significance   & **	&**	&-	&***	&*	&**	&***	&***	&**	&***	&***	&***\\
\bottomrule
\end{tabular*}
\end{center}
\end{table*}

\begin{figure}[t]
    \centering
    \subfloat[Oncology-Screen\_CV2]{\includegraphics[width=0.45\linewidth]{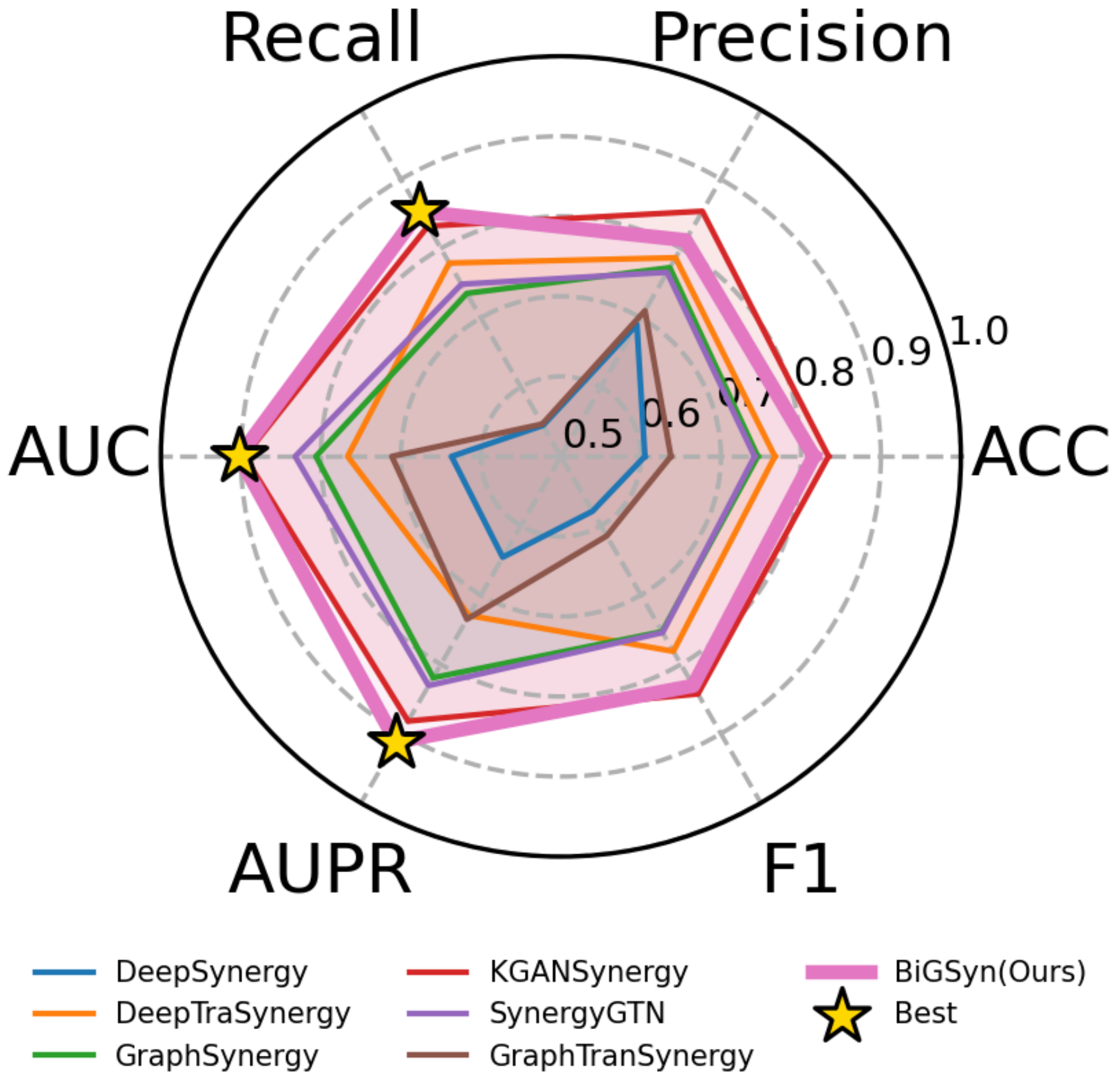}\label{fig:Oneil_CV2}}
    \hfil
    \subfloat[DrugCombDB\_CV2]{\includegraphics[width=0.45\linewidth]{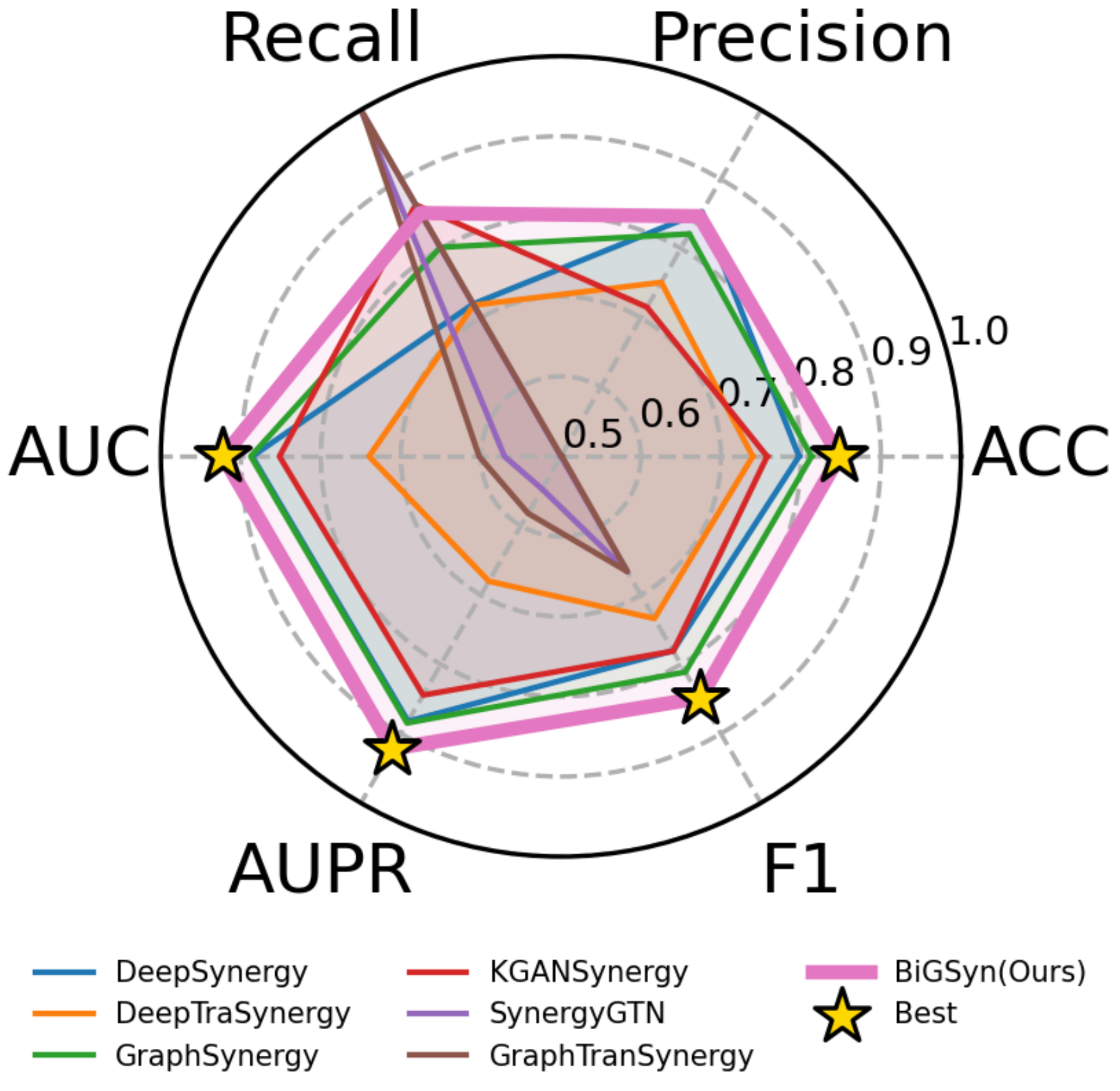}\label{fig:DrugCombDB_CV2}}
    \\
    \subfloat[Oncology-Screen\_CV3]{\includegraphics[width=0.45\linewidth]{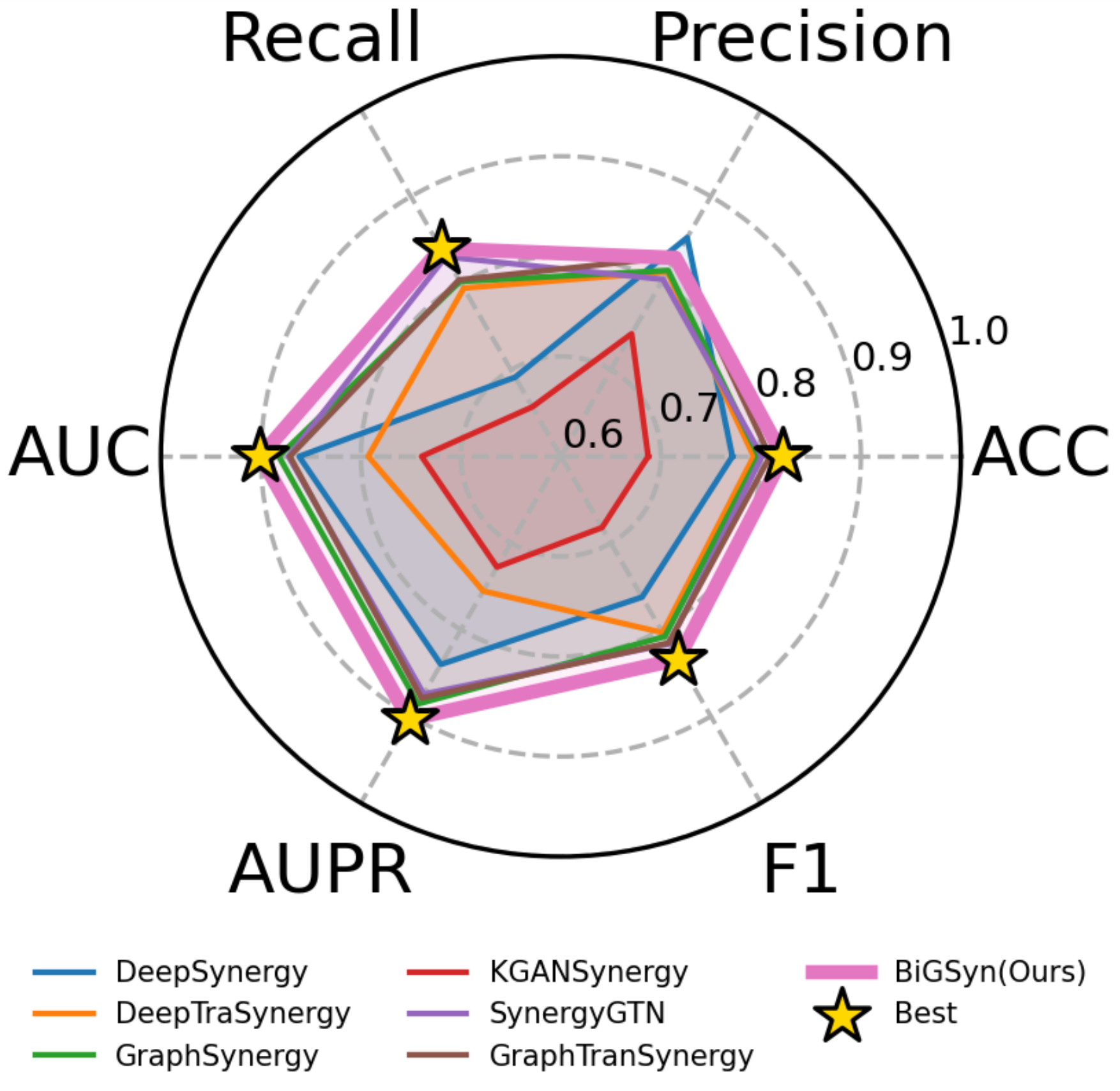}\label{fig:Oneil_CV3}}
    \hfil
    \subfloat[DrugCombDB\_CV3]{\includegraphics[width=0.45\linewidth]{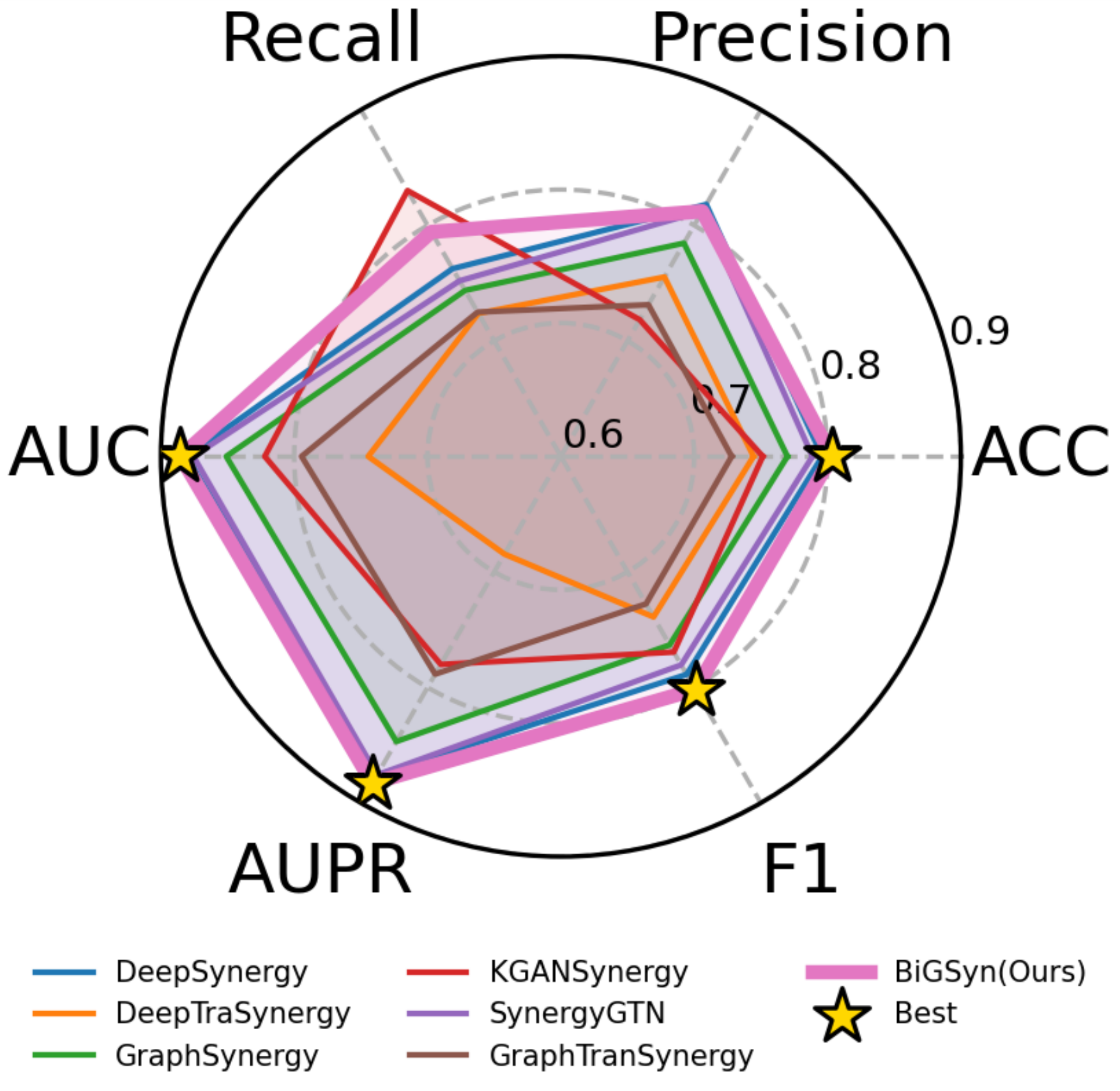}\label{fig:DrugCombDB_CV3}}
    \caption{Test performance comparison of different models on two datasets under CV2 and CV3 settings.}
    \label{fig:CV2_CV3_radar}
\end{figure}

\begin{figure}[t]
    \centering
    \subfloat[Oncology-Screen\_CV2]{\includegraphics[width=0.43\linewidth]{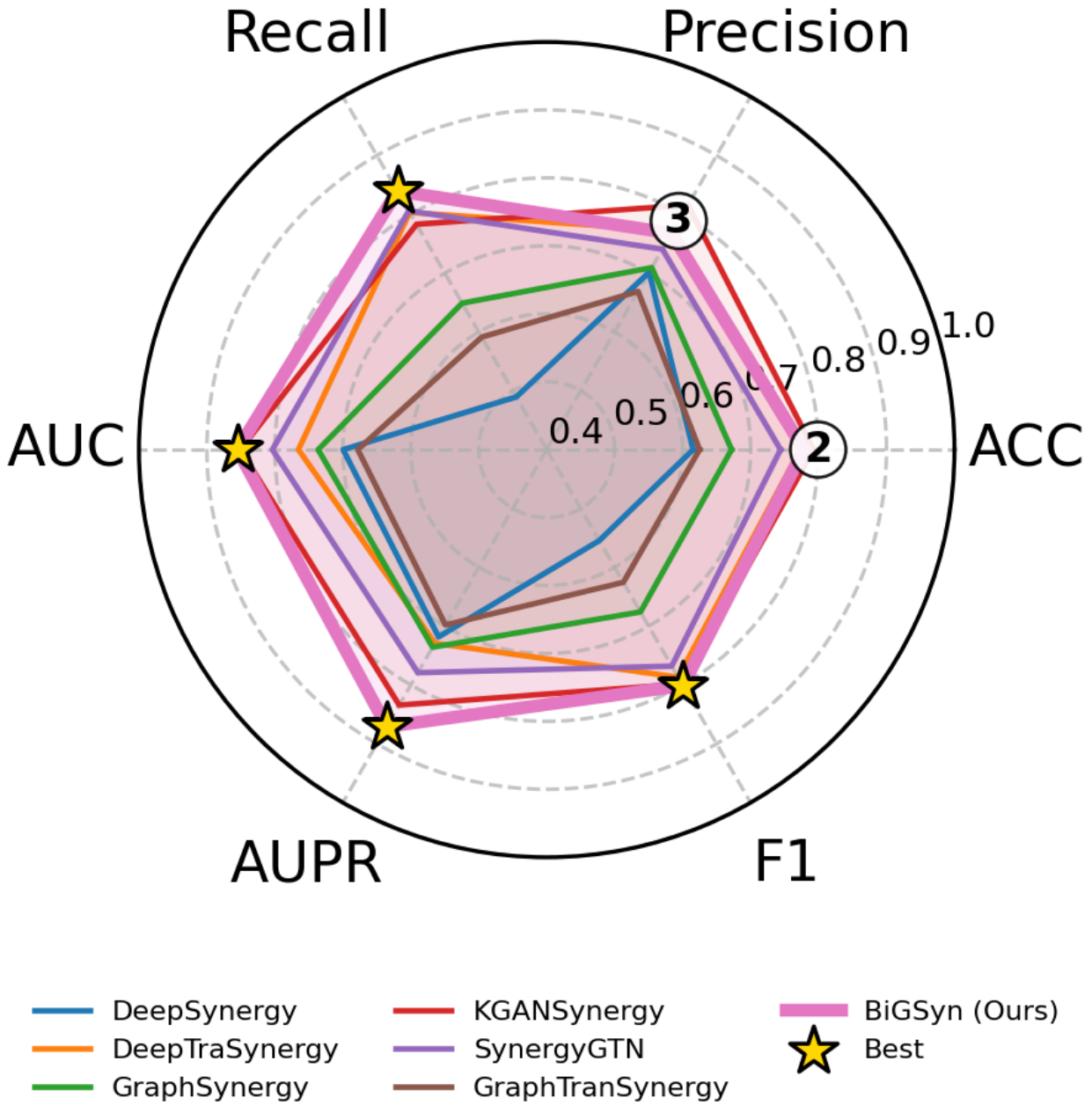}\label{fig:Oneil_CV2_cv}}
    \hfil
    \subfloat[DrugCombDB\_CV2]
    {\includegraphics[width=0.43\linewidth]{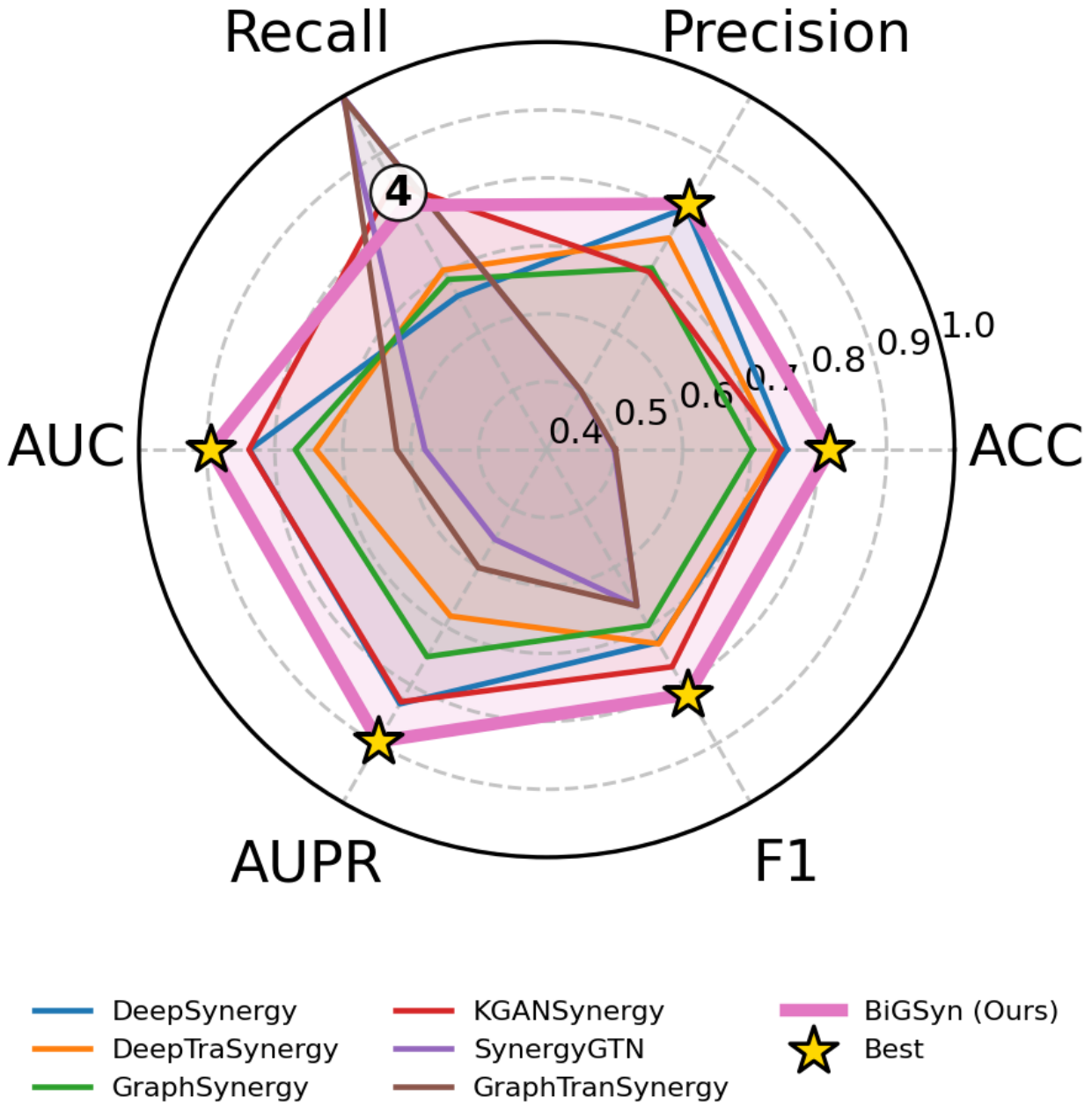}\label{fig:DrugCombDB_CV2_cv}}
    \\
    \subfloat[Oncology-Screen\_CV3]{\includegraphics[width=0.43\linewidth]{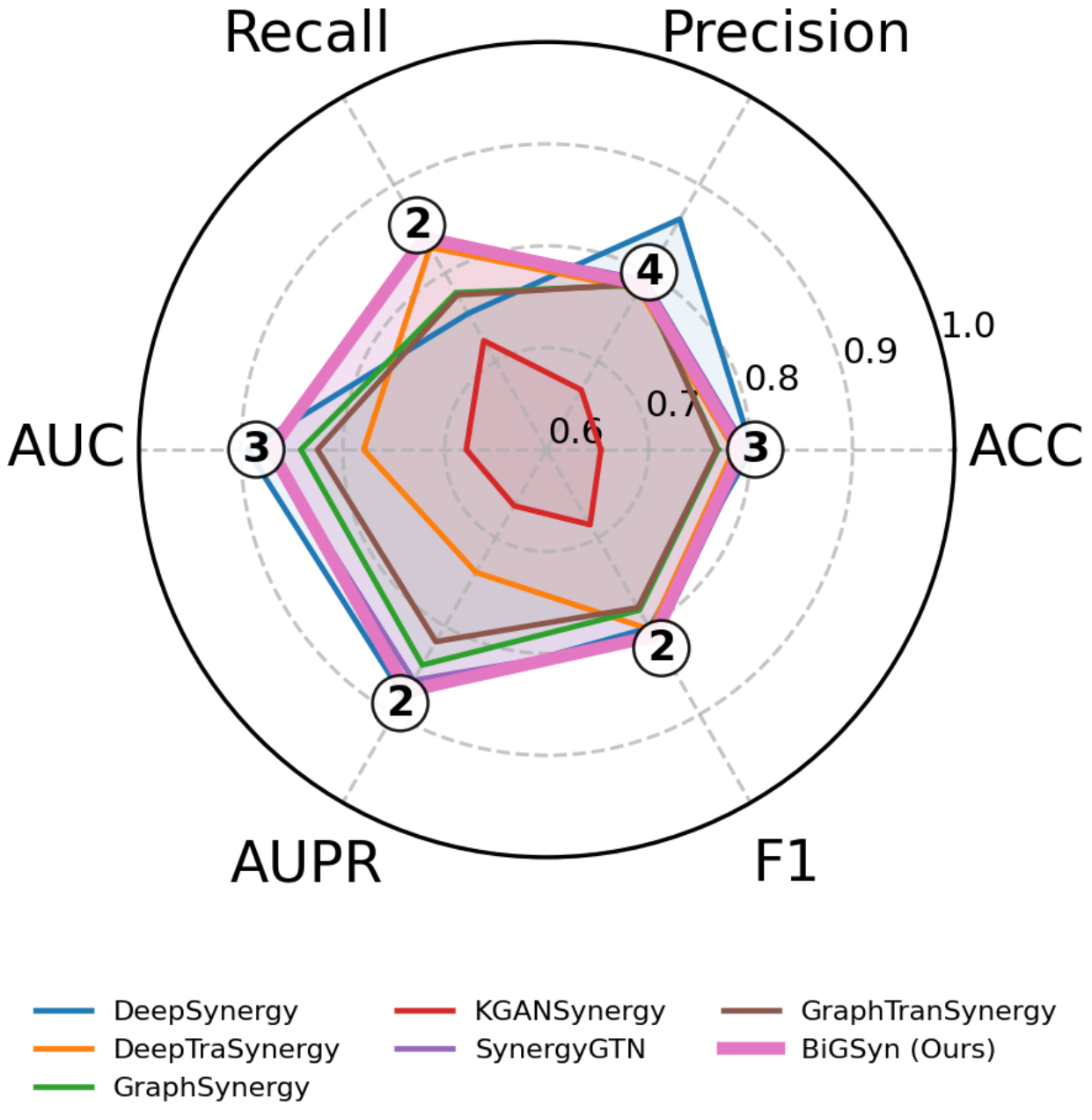}\label{fig:Oneil_CV3_cv}}
    \hfil
    \subfloat[DrugCombDB\_CV3]
    {\includegraphics[width=0.43\linewidth]{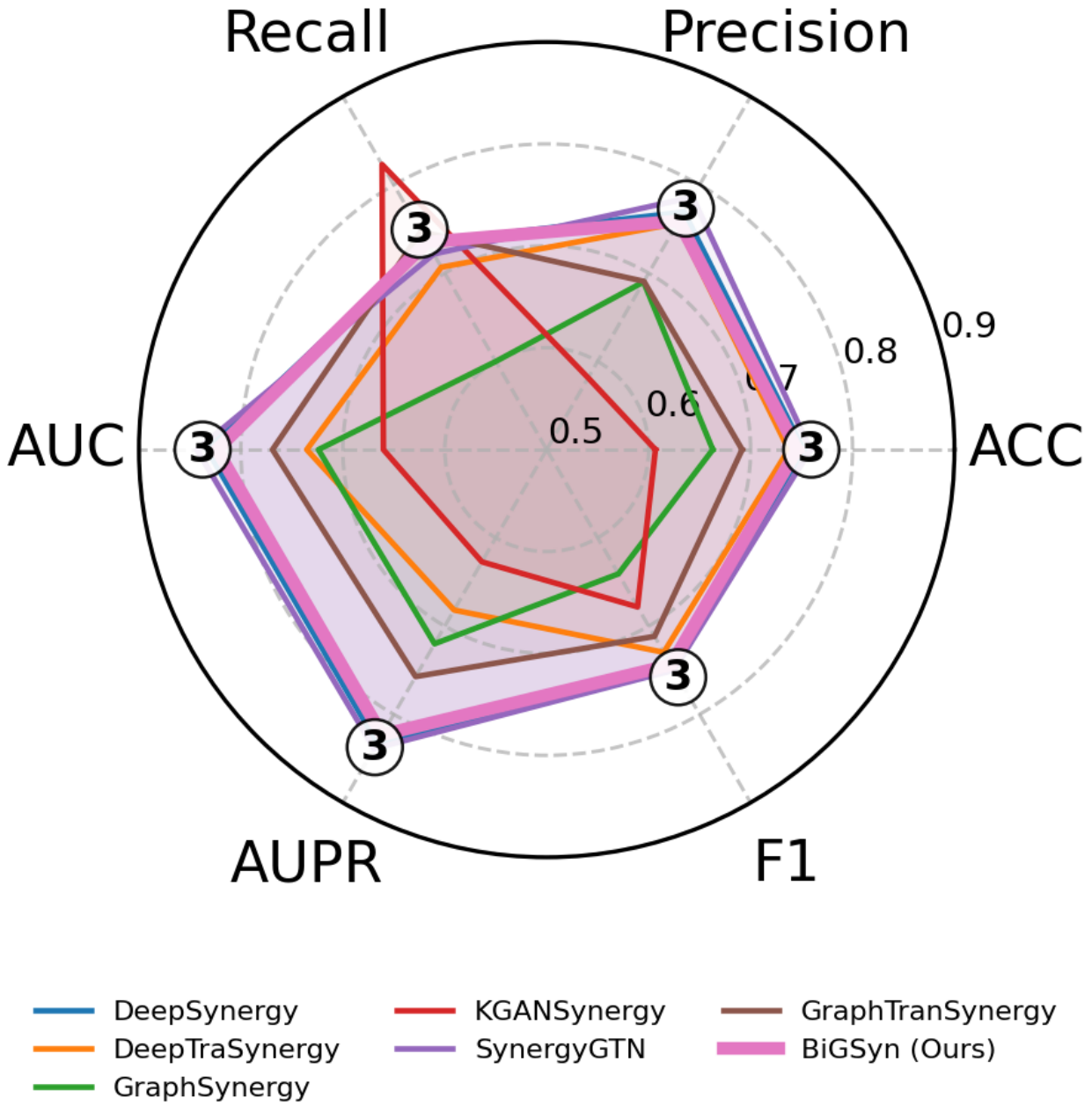}\label{fig:DrugCombDB_CV3_cv}}
    \caption{CV results of different models on the Oncology-Screen and DrugCombDB datasets under CV2 and CV3 settings.}
    \label{fig:CV2_CV3_radar_cv}
\end{figure}

Table~\ref{tab:CV1} presents the independent test results under the CV1 setting on both benchmark datasets, evaluated using six primary metrics: ACC, Precision, Recall, AUC, AUPR, and F1-score. 
Test results for CV2 and CV3 settings are provided in Fig.~\ref{fig:CV2_CV3_radar}
The main observations are as follows: 
(1) Under the CV1 setting, graph-based approaches exhibit superior predictive performance, particularly on the small-scale Oncology-Screen dataset. 
Our proposed BiGSyn achieves the best results across all metrics on both datasets. 
(2) Under CV2 training, where drug-pair-associated synergy samples in validation folds are excluded from training, molecular-graph-only methods such as SynergyGTN and GraphTranSynergy show unstable independent test performance. 
(3) In the CV3 scenario, where cell-line-associated synergy samples in validation folds are excluded from training, most baselines show marked performance drops. Nevertheless, BiGSyn maintains the best overall results, significantly outperforming other models in terms of ACC, AUC, AUPR and F1-score. 
(4) Overall, BiGSyn consistently delivers superior independent test performance after training under all three CV settings, demonstrating robustness across different synergy-label partitioning scenarios. 

In the 5-fold CV results shown in Fig.~\ref{fig:CV2_CV3_radar_cv}, BiGSyn consistently achieves strong performance under CV2, demonstrating robust prediction for synergy samples associated with held-out drug pairs. 
Under CV3, validation performance is slightly lower, reflecting the increased difficulty of predicting synergy labels associated with held-out cell lines. 
Nevertheless, BiGSyn remains competitive in CV3 and achieves the best independent test performance across multiple metrics.

\subsection{Ablation Studies}

\begin{figure}[t]  
    \centering
    \includegraphics[width=0.8\linewidth]{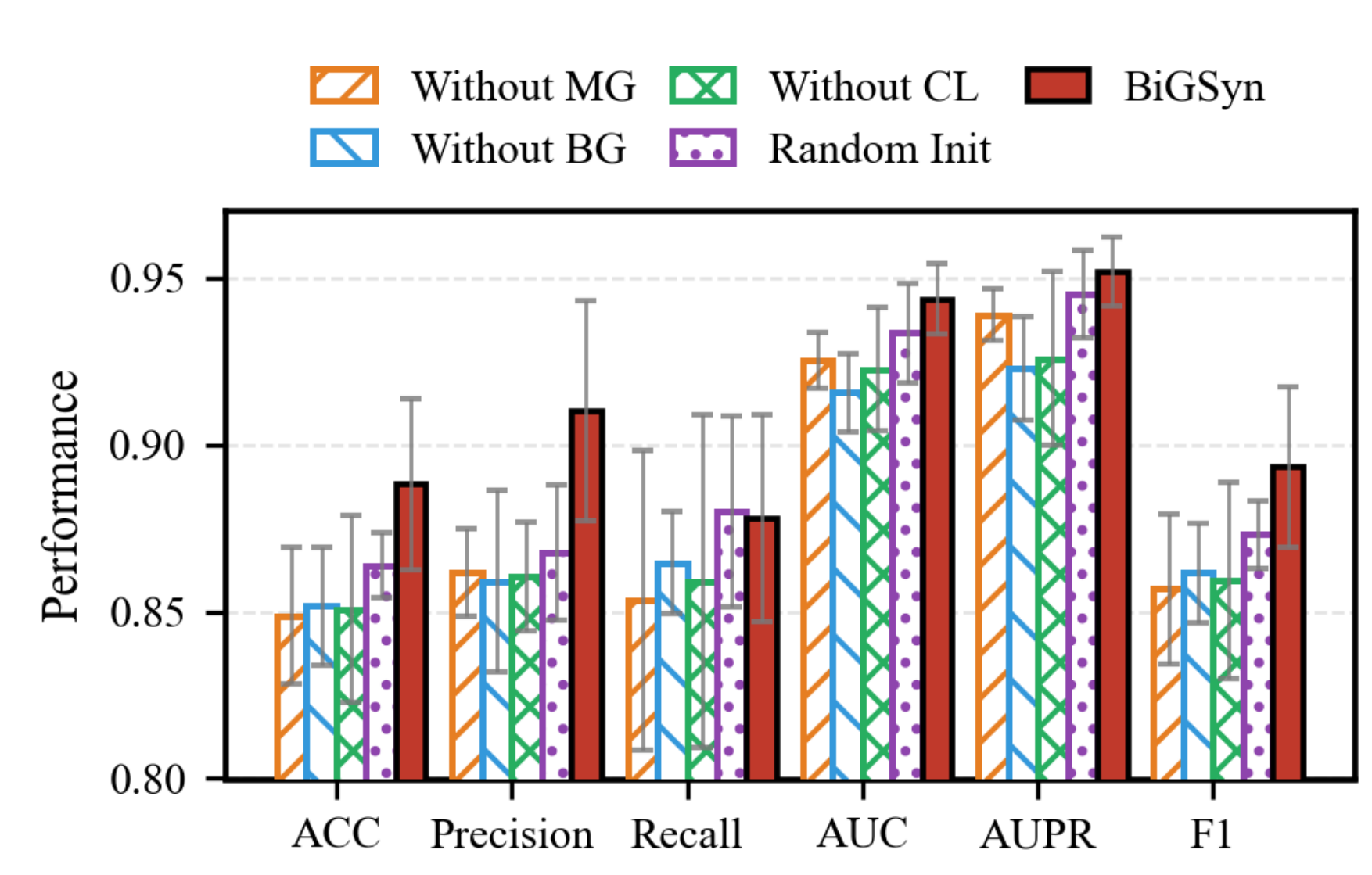}\label{fig_MAS_OS}
    \caption{Performance comparison between different variants on the Oncology-Screen dataset under the CV1 scenario.}
    \label{fig_MAS1}
\end{figure}

To investigate the contribution of each component in BiGSyn, we designed four ablation variants by (i) removing the molecular graph-level representation (Without MG) to assess the role of the structural information, (ii) removing the biological graph-level representation (Without BG) to evaluate the role of the contextual information, (iii) removing the contrastive learning module (Without CL) to measure the effectiveness of bi-level alignment, and (iv) randomly initializing nodes instead of using sequence-based features (Random Init) to assess the impact of semantic enhancement in node representations. 
As shown in Fig.~\ref{fig_MAS1}, removing any component leads to a clear performance drop. 
Removing structural modality information (Without MG) causes the most significant performance degradation across all evaluation metrics, indicating that structural information forms the foundation for effective drug representation. 
Eliminating contextual modality information (Without BG) also leads to consistent performance declines, highlighting the importance of modeling biological interactions. 
Replacing sequence-based initialization with random embeddings (Random Init) also reduces performance, confirming that sequence-level semantic information provides meaningful biological priors. 
Removing the contrastive learning module (Without CL) results in additional performance drops, demonstrating the effectiveness of bi-level alignment in improving representation quality. 
Overall, the results validate the effectiveness of jointly leveraging complementary structural, contextual, and sequential modalities, together with bi-level alignment via contrastive learning, to enhance predictive performance. 

\begin{figure}[t]  
    \centering
    \subfloat[BiGSyn]{\includegraphics[width=0.42\linewidth]{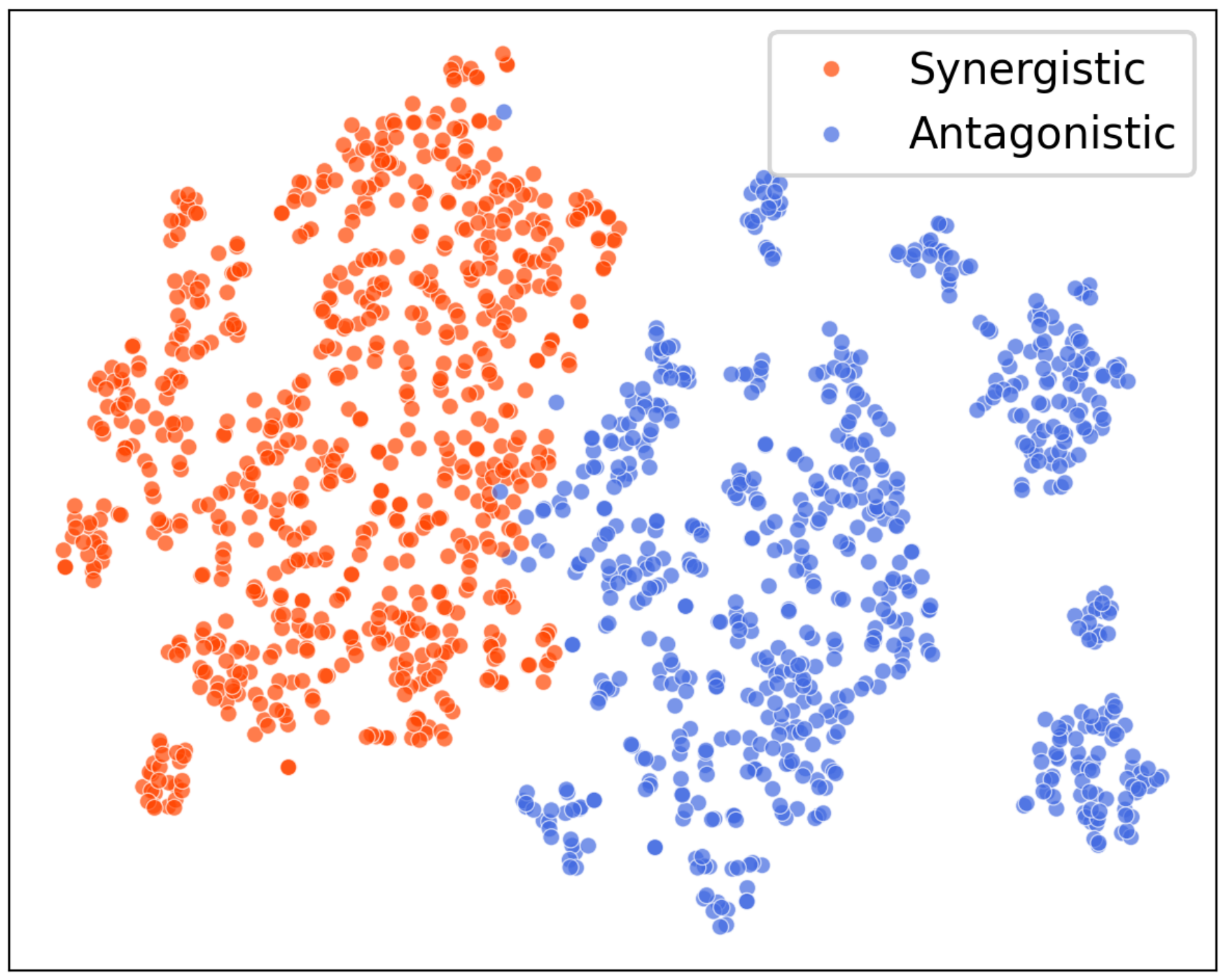}%
    \label{fig_MAS_both}}
    \hfil
    \subfloat[Without MG]{\includegraphics[width=0.42\linewidth]{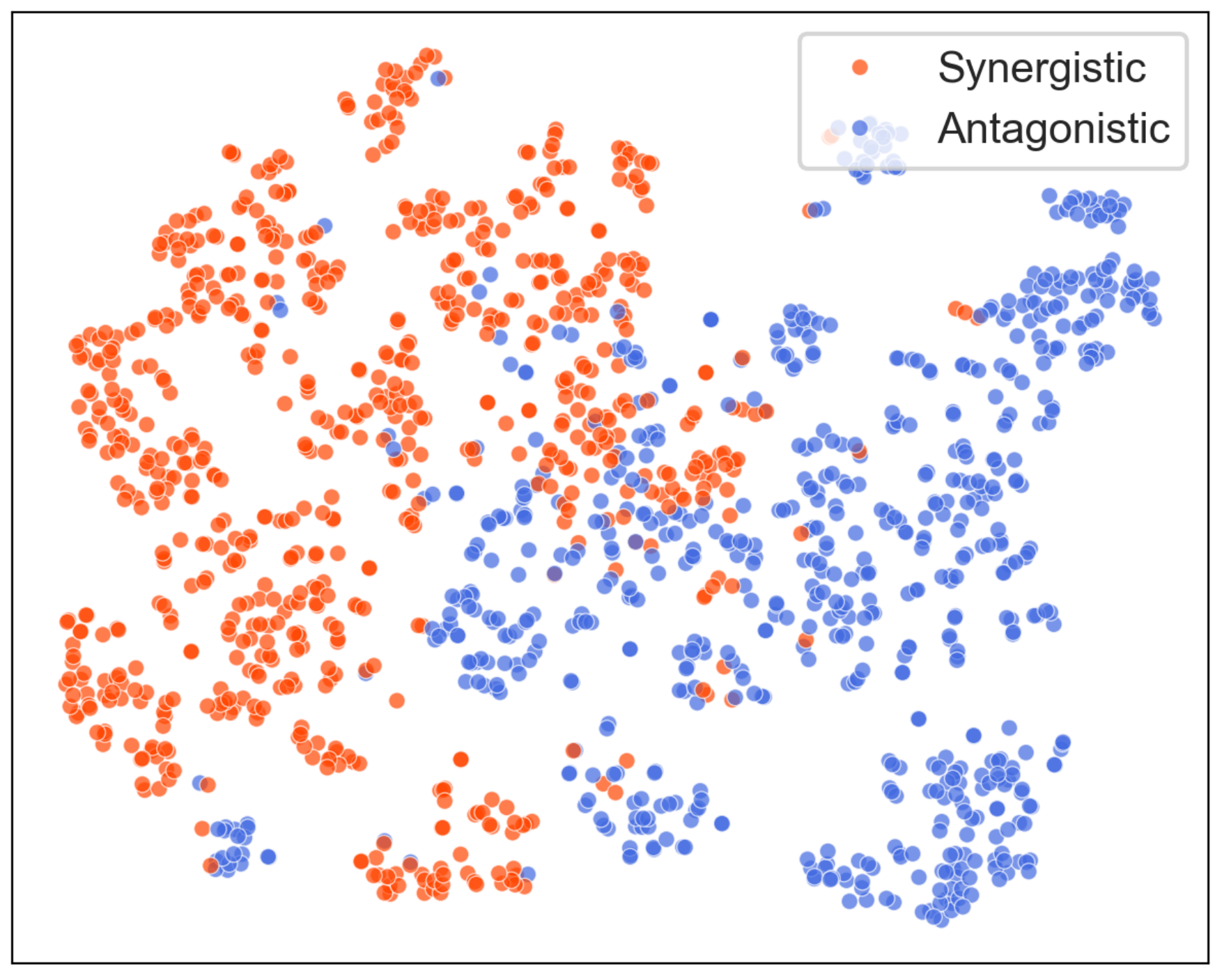}%
    \label{fig_MAS_rel}}
    \\
    \subfloat[Without BG]{\includegraphics[width=0.42\linewidth]{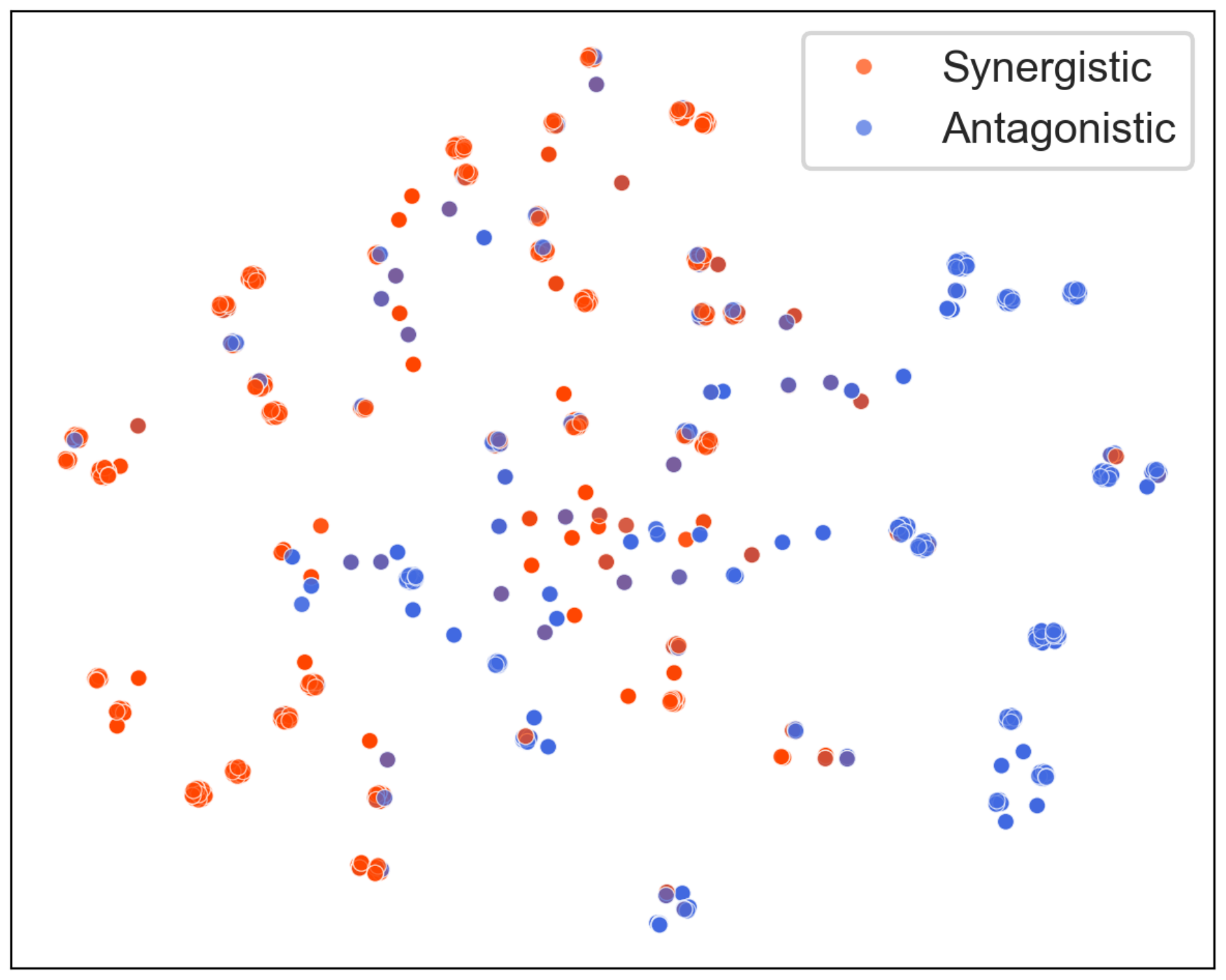}%
    \label{fig_MAS_mol}}
    \hfil
    \subfloat[Without CL]{\includegraphics[width=0.42\linewidth]{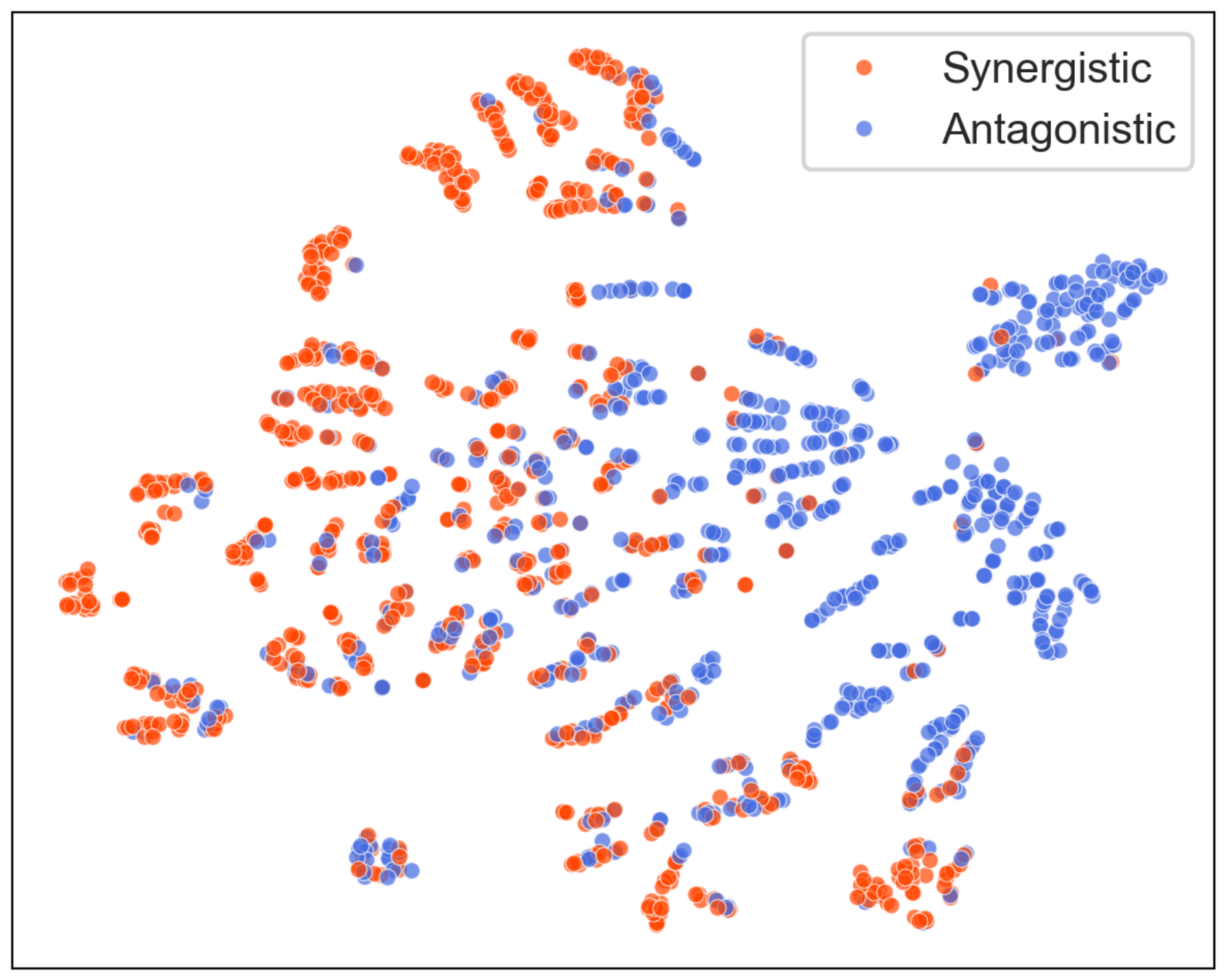}%
    \label{fig_MAS_cl}}
    \caption{t-SNE visualization of the embeddings from the last layer.}
    \label{fig_MAS2}
\end{figure}

To further assess the effectiveness of the bi-level representation, we applied t-SNE to visualize the embeddings of drug–drug–cell line triples generated by the last layer of BiGSyn. As shown in Fig.~\ref{fig_MAS_both}, the bi-level embeddings achieve a clear separation between synergistic and antagonistic triples, demonstrating the discriminative power of the learned representation. 
In contrast, when only the biological graph is used (Fig.~\ref{fig_MAS_rel}), the two classes become less distinguishable, with evident overlaps between clusters. 
The molecular graph alone (Fig.~\ref{fig_MAS_mol}) leads to even more severe mixing, where synergistic and antagonistic triples are heavily entangled. 
Moreover, when contrastive learning is removed (Fig.~\ref{fig_MAS_cl}), the embeddings lose their structural compactness, resulting in scattered distributions and blurred class boundaries, indicating that contrastive learning plays a crucial role in enhancing intra-class cohesion and inter-class separability. 
These observations confirm that the integration of multiple levels enables BiGSyn to capture complementary information and generate more meaningful low-dimensional embeddings related to drug synergy.

\subsection{Hyper-parameter Sensitivity Analysis}\label{sec:param}

\begin{figure}[t]
    \centering
    \subfloat[]{\includegraphics[width=0.46\linewidth]{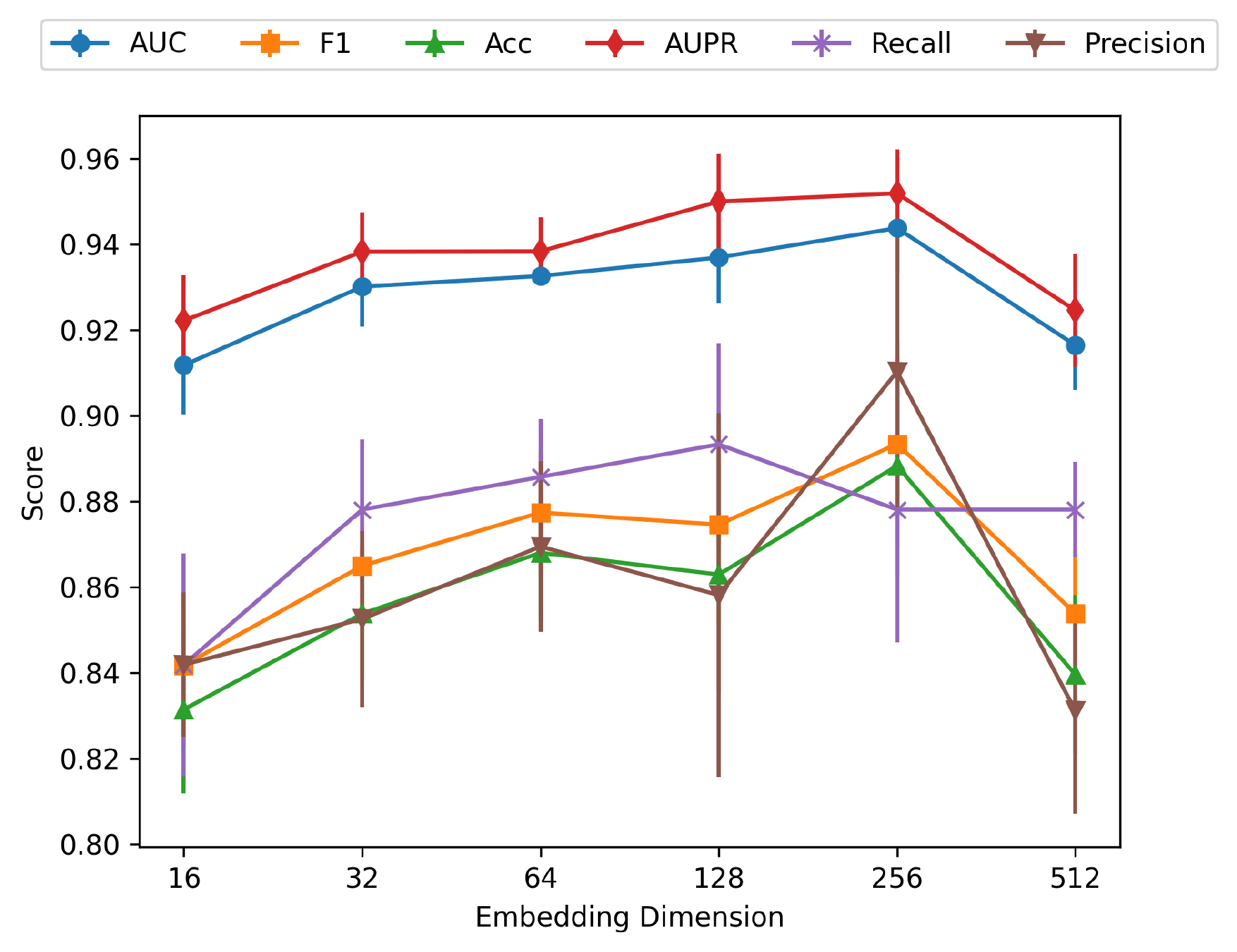}\label{fig_PSA_d}}
    \hfil
    \subfloat[]{\includegraphics[width=0.46\linewidth]{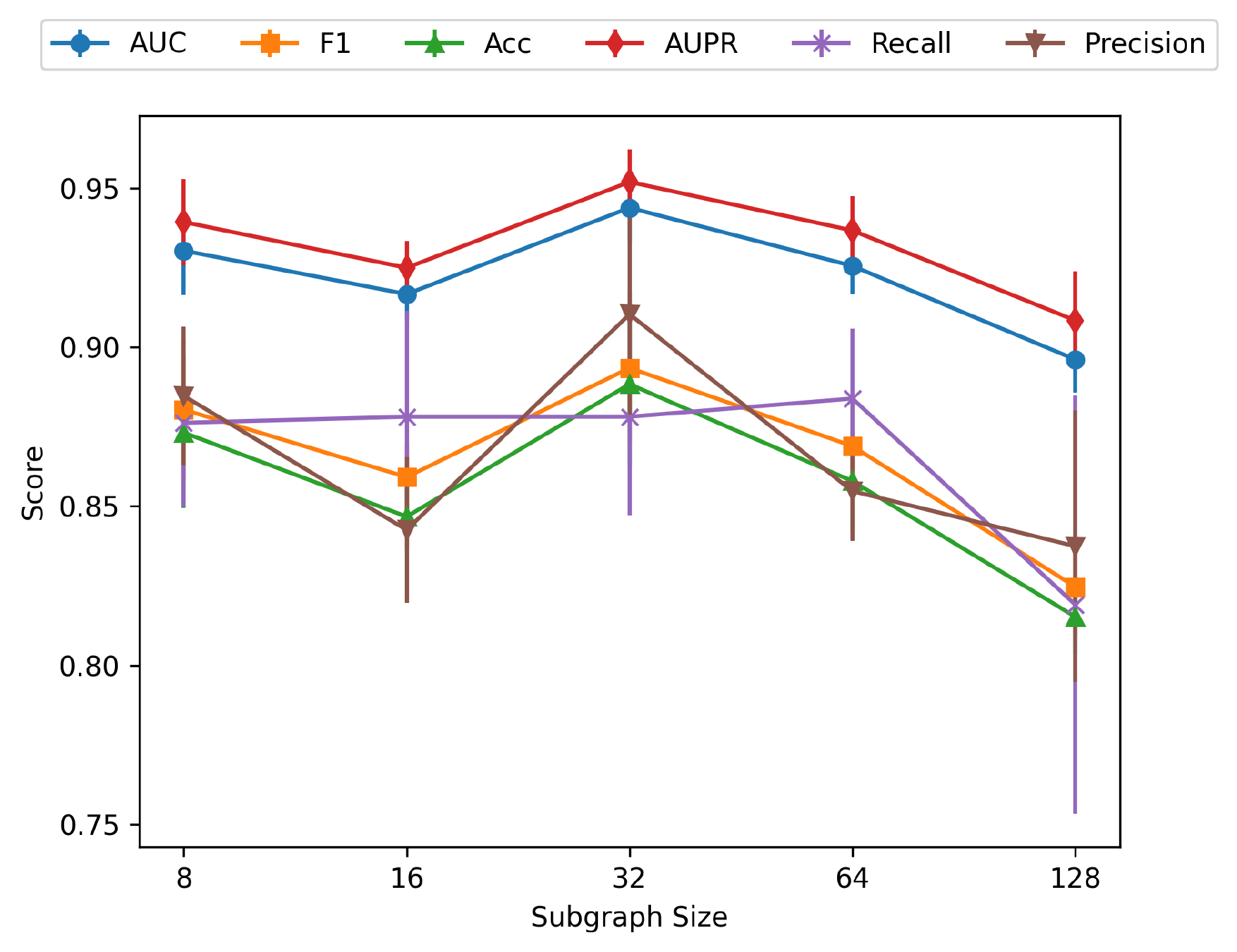}\label{fig_PSA_s}}
    
    \vspace{-3mm}  
    \subfloat[]{\includegraphics[width=0.46\linewidth]{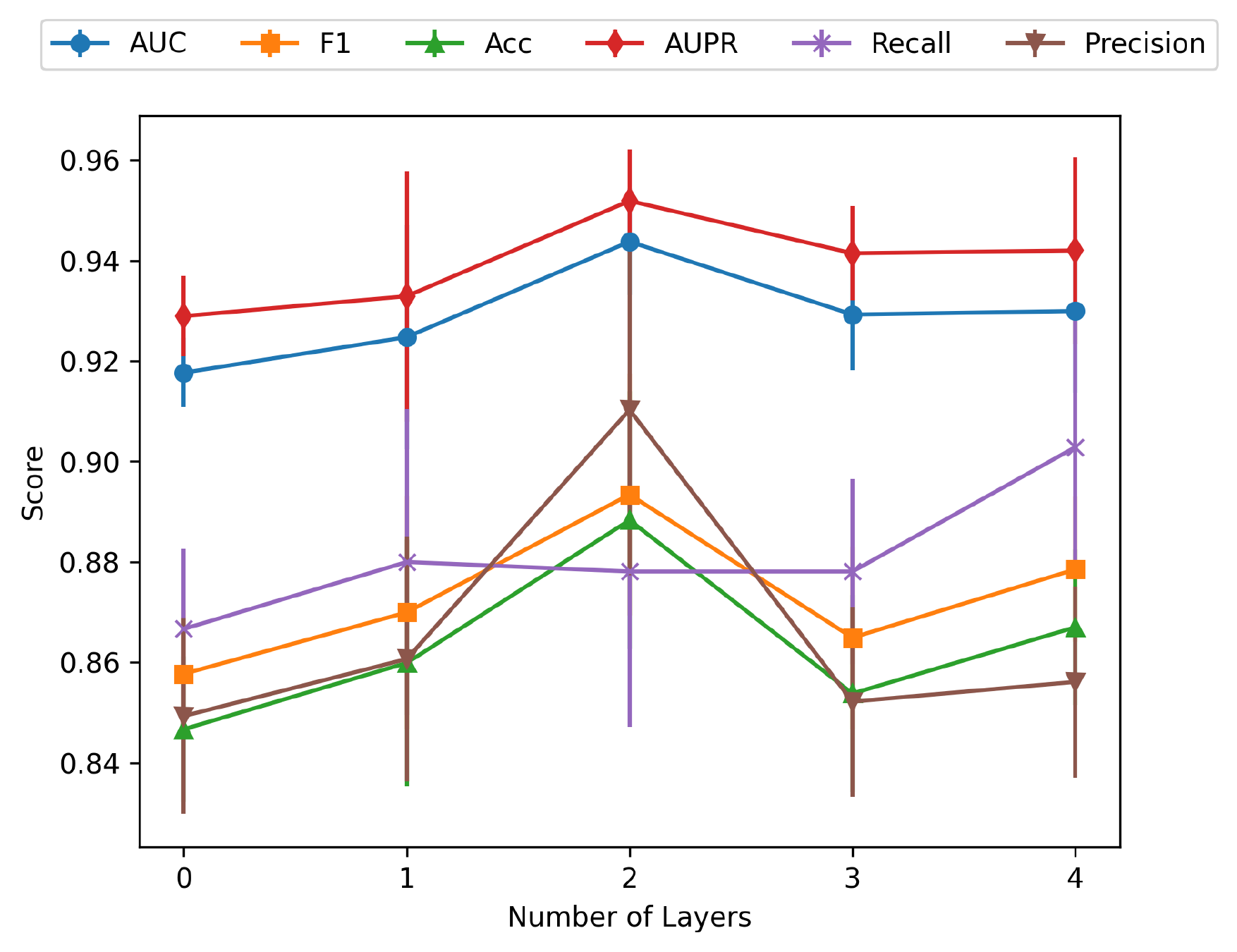}\label{fig_PSA_l}}
    \hfil
    \subfloat[]{\includegraphics[width=0.46\linewidth]{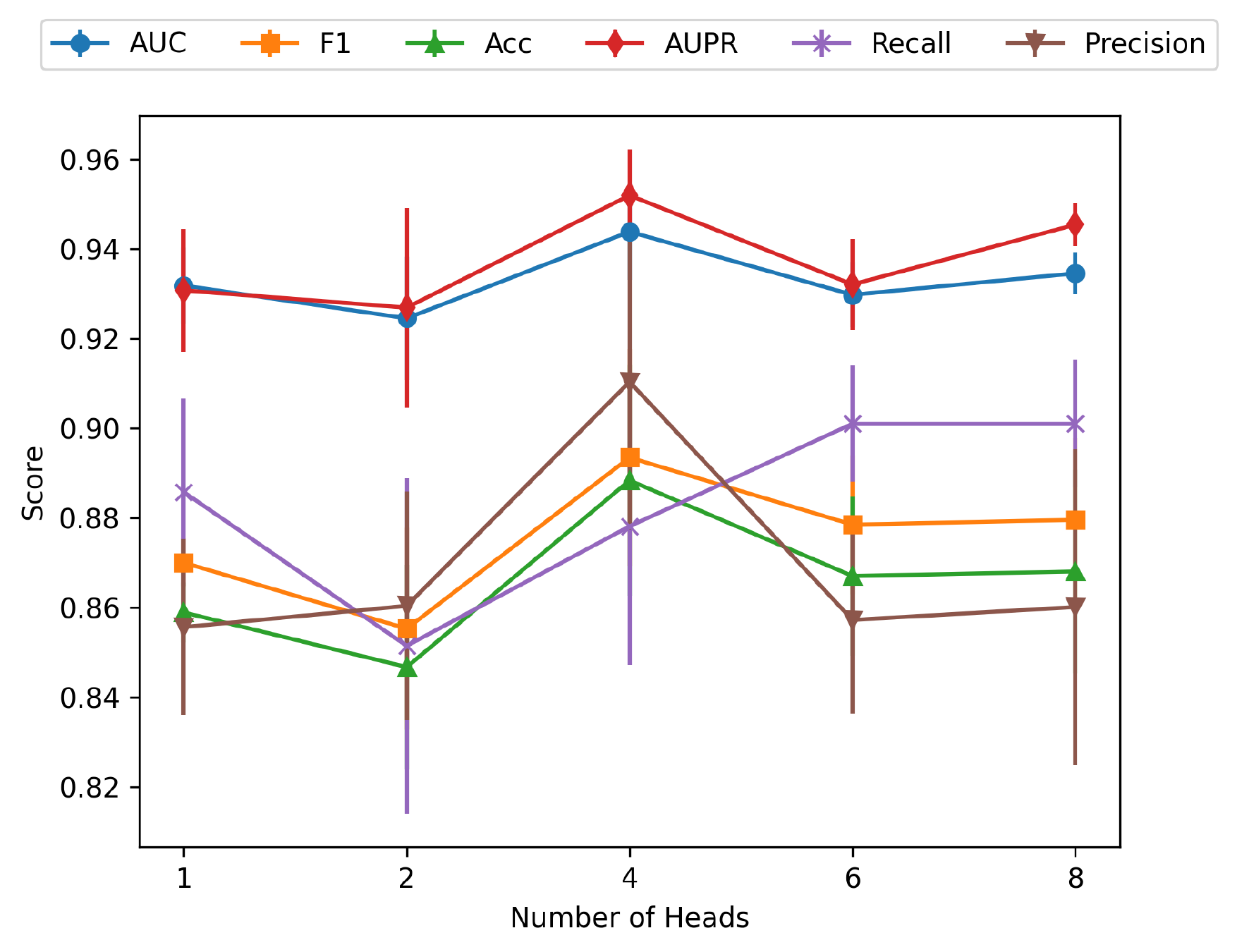}\label{fig_PSA_h}}
    
    \caption{Parameter sensitivity analysis for BiGSyn.}
    \label{fig_PSA}
\end{figure}

We conducted a sensitivity analysis on four key hyperparameters, including embedding dimension, subgraph size, number of layers, and number of attention heads. As shown in Fig.~\ref{fig_PSA}, BiGSyn demonstrates stable performance across different hyperparameter settings, with identifiable optimal configurations. Increasing the embedding dimension generally improves performance until it peaks at 256, after which overfitting slightly reduces accuracy. 
The subgraph size controls the locality--coverage trade-off in the biological network, where smaller subgraphs focus on the local neighborhoods, whereas larger subgraphs incorporate more distant PPI and tissue-mediated paths but may introduce hub-dominated relations and task-irrelevant noise. A subgraph size of 32 achieves the best performance in this case. 
In terms of model depth, two layers yield the highest scores, while deeper architectures suffer from over-smoothing. Similarly, multi-head attention benefits the model, with four heads providing the best trade-off, whereas further increasing the number of heads brings limited or unstable gains. Overall, these results indicate that BiGSyn is robust to parameter variations, and its optimal configuration is achieved with an embedding dimension of 256, subgraph size of 32, two layers, and four attention heads. 

\begin{table*}[t]
\caption{Validation for predicted unknown drug combinations on the Oncology-Screen dataset.\label{tab:case}}%
\begin{center}
\begin{tabular*}{\textwidth}{@{\extracolsep\fill}lrrrrrr@{\extracolsep\fill}}  
\toprule
Cancer      & Cell line & Drug A & Drug B & Predicted probability & PMID\\
\midrule
Ovarian     & A2780 & Paclitaxel & Vinorelbine & 0.9954257    & 21528349 \\
colorectal  & HT29  & Paclitaxel & Vinorelbine & 0.9839717    & 9625431  \\
Lung        & NCI-H460 & Sorafenib  & Vinorelbine & 0.9719465 & 16724239 \\
Ovarian     & SKOV3 & Dasatinib  & Topotecan   & 0.9450858    & 24407585 \\
Lung        & NCI-H23 & Erlotinib   & Paclitaxel & 0.9399144  & 29536182 \\
\bottomrule
\end{tabular*}
\end{center}
\end{table*}

\subsection{Discovering Novel Drug Combinations}\label{sec:case}

For the purpose of predicting novel synergistic drug combinations, we further conduct case studies on the Oncology-Screen dataset. 
Among all 6,090 possible drug–drug–cell line triplets derived from 21 drugs and 29 cell lines, 67\% (4,130) triplets do not have experimentally measured synergy scores. 
In this setting, all confirmed triplets with available synergy measurements were used as training data, while the remaining unmeasured triplets were treated as candidate samples. 
After training, BiGSyn generated prediction scores for the candidate triplets, which were subsequently ranked according to their predicted values. 
To evaluate the reliability of these predictions, we performed a non-exhaustive literature search. 
Several top-ranked candidate triplets were supported by independent experimental evidence reported in previous studies. 
The PubMed identifiers (PMIDs) of representative examples are summarized in Table~\ref{tab:case}. 
These results indicate that BiGSyn can effectively prioritize true synergistic drug pairs from numerous candidates and identify potential synergies beyond the scope of existing training data. 

\subsection{Interpretability of BiGSyn}

We take the predicted synergistic combination of Paclitaxel and Vinorelbine on A2780 in Section~\ref{sec:case}
as a representative case. 
We further analyze the attention distributions on both the molecular graphs and the biological subgraphs, which reveal the key substructures and relations contributing to the prediction. 

\begin{figure}[t]  
    \centering
    \includegraphics[width=\linewidth]{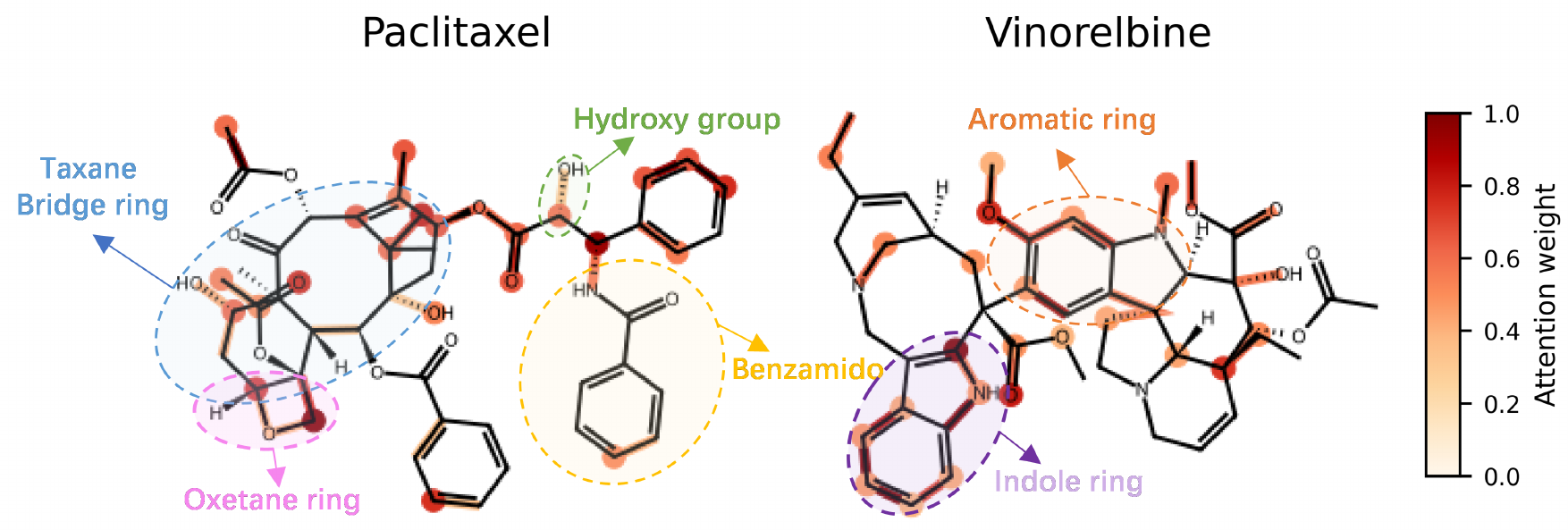}%
    \caption{The molecular attention maps.}
    \label{fig_mol_attn}
\end{figure}

In the visualized molecular attention maps of Paclitaxel and Vinorelbine (Fig.~\ref{fig_mol_attn}), the highlighted atoms and bonds with the top 40\% of attention scores are primarily located on the ester bonds, oxetane ring, and substituted phenyl groups of Paclitaxel, as well as the indole–vincamine core of Vinorelbine. 
These regions coincide with the pharmacologically critical motifs previously identified in structural and biochemical studies. 
Specifically, Paclitaxel’s C-13 ester side chain and oxetane ring have been shown to play essential roles in its binding to $\beta$-tubulin and microtubule stabilization~\cite{prota2023structural}.
The phenyl substituents on the taxane skeleton also significantly affect the binding affinity and microtubule stabilization potential~\cite{prota2023structural}. 
Similarly, the high-attention regions in Vinorelbine correspond to its vinca alkaloid backbone, where the aromatic and nitrogen-containing rings are known to mediate tubulin binding and inhibit microtubule polymerization~\cite{coderch2012tubulin}. 
These consistent patterns suggest that the attention mechanism effectively captures the structure–function relationships underlying microtubule-targeting activity, demonstrating the interpretability of the learned molecular representations in explaining drug synergy behavior. 

\begin{figure}[t]  
    \centering
    \includegraphics[width=\linewidth]{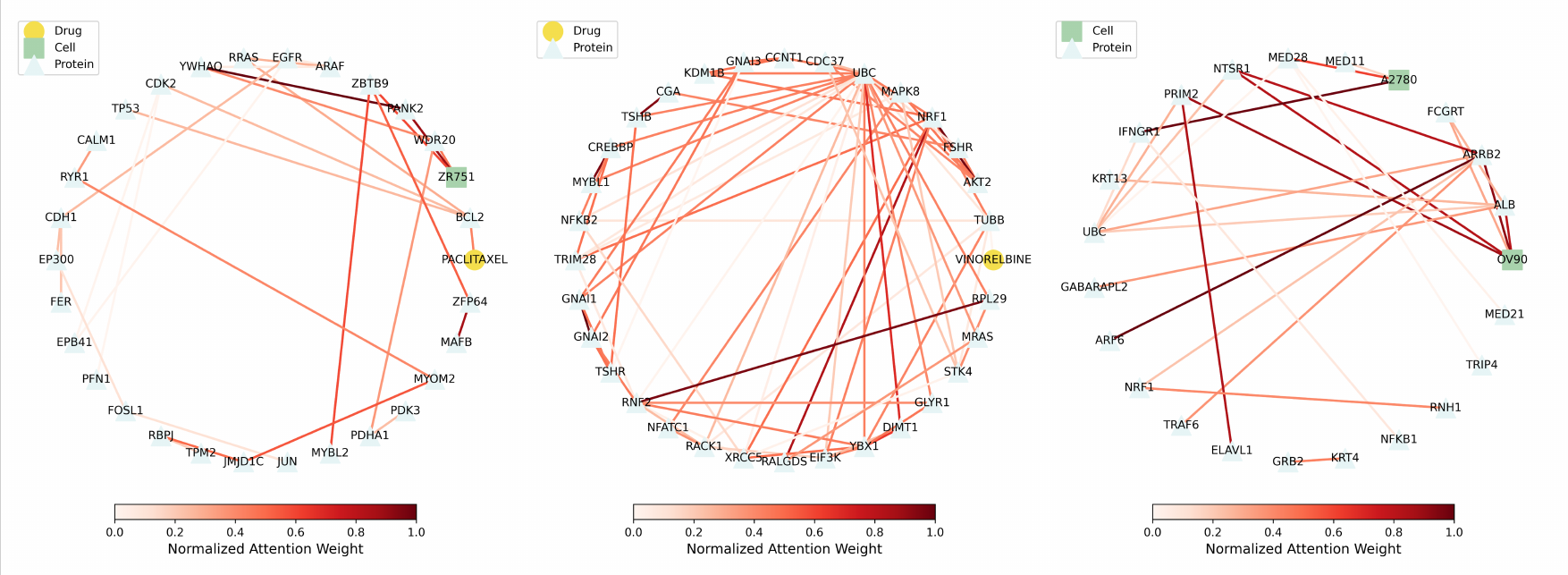}%
    \caption{Attention-based subgraphs highlighting key molecular and cellular interactions identified by the trained model.}  
    \label{fig_rel_attn}
\end{figure}

The attention distributions in the biological subgraphs of Paclitaxel, Vinorelbine, and A2780 cell line reveal distinct focal patterns, as shown in Fig.~\ref{fig_rel_attn}. 
In the Paclitaxel subgraph, protein nodes such as EGFR (epidermal growth factor receptor) are located in highly attended neighborhoods, consistent with Paclitaxel’s key mechanism of indirectly modulating EGFR endocytosis via microtubule dynamics~\cite{10.1371/journal.pone.0045465}. 
Meanwhile, UBC (Ubiquitin C) appears as a central hub with many high-weight edges in both the Vinorelbine and A2780 subgraphs, suggesting that ubiquitin-mediated proteostasis may act as a shared integrator of drug and cell responses, aligning with the established role of dysregulated ubiquitination in tumorigenesis and stress adaptation~\cite{deng2020role,duan2021ubiquitin}. 

\section{Conclusion}


In this work, we propose BiGSyn, a bi-level heterogeneous graph contrastive learning framework for drug synergy prediction. 
By using heterogeneous graph transformers to jointly model fine-grained structural patterns in molecular graphs and coarse-grained relational contexts in biological interaction networks, BiGSyn learns more comprehensive drug representations. 
A contrastive fusion mechanism further aligns representations across the two levels, bridging the semantic gap between chemical structures and biological contexts. 
Extensive experiments on two benchmark datasets demonstrate its superior predictive performance and strong generalization ability under challenging evaluation settings. 
Moreover, the attention mechanism provides interpretable insights by highlighting key molecular substructures and biological relations associated with drug synergy. 
These results underscore the value of cross-level representation learning for accurate and interpretable drug synergy prediction.

\section*{Acknowledgment}
This work was supported in part by the National Natural Science Foundation of China (72371148 and 72192832) and the Liaoning Provincial Youth Science Foundation (Category B; formerly the Provincial Excellent Young Scientists Fund) (2024L0351).

\bibliographystyle{IEEEtran}
\bibliography{main}

@article{lehar2009synergistic,
  title={Synergistic drug combinations tend to improve therapeutically relevant selectivity},
  author={Leh{\'a}r, Joseph and Krueger, Andrew S and Avery, William and Heilbut, Adrian M and Johansen, Lisa M and Price, E Roydon and Rickles, Richard J and Short Iii, Glenn F and Staunton, Jane E and Jin, Xiaowei and others},
  journal={Nature Biotechnology},
  volume={27},
  number={7},
  pages={659--666},
  year={2009},
  publisher={Nature Publishing Group US New York}
}

@article{jaaks2022effective,
  title={Effective drug combinations in breast, colon and pancreatic cancer cells},
  author={Jaaks, Patricia and Coker, Elizabeth A and Vis, Daniel J and Edwards, Olivia and Carpenter, Emma F and Leto, Simonetta M and Dwane, Lisa and Sassi, Francesco and Lightfoot, Howard and Barthorpe, Syd and others},
  journal={Nature},
  volume={603},
  number={7899},
  pages={166--173},
  year={2022},
  publisher={Nature Publishing Group UK London}
}

@article{jin2023rational,
  title={Rational combinations of targeted cancer therapies: background, advances and challenges},
  author={Jin, Haojie and Wang, Liqin and Bernards, Ren{\'e}},
  journal={Nature Reviews Drug Discovery},
  volume={22},
  number={3},
  pages={213--234},
  year={2023},
  publisher={Nature Publishing Group UK London}
}

@article{abbasi2024new,
  title={New methods for drug synergy prediction: A mini-review},
  author={Abbasi, Fatemeh and Rousu, Juho},
  journal={Current Opinion in Structural Biology},
  volume={86},
  pages={102827},
  year={2024},
  publisher={Elsevier}
}

@article{pourmousa2025ai,
  title={AI-driven discovery of synergistic drug combinations against pancreatic cancer},
  author={Pourmousa, Mohsen and Jain, Sankalp and Barnaeva, Elena and Jin, Wengong and Hochuli, Joshua and Itkin, Zina and Maxfield, Travis and Melo-Filho, Cleber and Thieme, Andrew and Wilson, Kelli and others},
  journal={Nature Communications},
  volume={16},
  number={1},
  pages={4020},
  year={2025},
  publisher={Nature Publishing Group UK London}
}

@article{sun2013high,
  title={High-throughput methods for combinatorial drug discovery},
  author={Sun, Xiaochen and Vilar, Santiago and Tatonetti, Nicholas P},
  journal={Science Translational Medicine},
  volume={5},
  number={205},
  pages={205rv1--205rv1},
  year={2013},
  publisher={American Association for the Advancement of Science}
}

@article{han2017synergistic,
  title={Synergistic drug combinations for cancer identified in a CRISPR screen for pairwise genetic interactions},
  author={Han, Kyuho and Jeng, Edwin E and Hess, Gaelen T and Morgens, David W and Li, Amy and Bassik, Michael C},
  journal={Nature Biotechnology},
  volume={35},
  number={5},
  pages={463--474},
  year={2017},
  publisher={Nature Publishing Group US New York}
}

@article{wang2022deepdds,
  title={DeepDDS: deep graph neural network with attention mechanism to predict synergistic drug combinations},
  author={Wang, Jinxian and Liu, Xuejun and Shen, Siyuan and Deng, Lei and Liu, Hui},
  journal={Briefings in Bioinformatics},
  volume={23},
  number={1},
  pages={bbab390},
  year={2022},
  publisher={Oxford University Press}
}

@article{besharatifard2024review,
  title={A review on graph neural networks for predicting synergistic drug combinations},
  author={Besharatifard, Milad and Vafaee, Fatemeh},
  journal={Artificial Intelligence Review},
  volume={57},
  number={3},
  pages={49},
  year={2024},
  publisher={Springer}
}

@article{xu2023dffndds,
  title={DFFNDDS: prediction of synergistic drug combinations with dual feature fusion networks},
  author={Xu, Mengdie and Zhao, Xinwei and Wang, Jingyu and Feng, Wei and Wen, Naifeng and Wang, Chunyu and Wang, Junjie and Liu, Yun and Zhao, Lingling},
  journal={Journal of Cheminformatics},
  volume={15},
  number={1},
  pages={33},
  year={2023},
  publisher={Springer}
}

@article{mao2016chemical,
  title={Chemical structure-related drug-like criteria of global approved drugs},
  author={Mao, Fei and Ni, Wei and Xu, Xiang and Wang, Hui and Wang, Jing and Ji, Min and Li, Jian},
  journal={Molecules},
  volume={21},
  number={1},
  pages={75},
  year={2016},
  publisher={MDPI}
}

@article{zhao2024molecular,
  title={Molecular property prediction based on graph structure learning},
  author={Zhao, Bangyi and Xu, Weixia and Guan, Jihong and Zhou, Shuigeng},
  journal={Bioinformatics},
  volume={40},
  number={5},
  pages={btae304},
  year={2024},
  publisher={Oxford University Press}
}

@article{nguyen2021graph,
  title={Graph convolutional networks for drug response prediction},
  author={Nguyen, Tuan and Nguyen, Giang TT and Nguyen, Thin and Le, Duc-Hau},
  journal={IEEE/ACM Transactions on Computational Biology and Bioinformatics},
  volume={19},
  number={1},
  pages={146--154},
  year={2021},
  publisher={IEEE}
}

@article{preuer2018deepsynergy,
  title={DeepSynergy: predicting anti-cancer drug synergy with deep learning},
  author={Preuer, Kristina and Lewis, Richard PI and Hochreiter, Sepp and Bender, Andreas and Bulusu, Krishna C and Klambauer, G{\"u}nter},
  journal={Bioinformatics},
  volume={34},
  number={9},
  pages={1538--1546},
  year={2018},
  publisher={Oxford University Press}
}

@article{kuru2021matchmaker,
  title={MatchMaker: a deep learning framework for drug synergy prediction},
  author={Kuru, Halil Ibrahim and Tastan, Oznur and Cicek, A Ercument},
  journal={IEEE/ACM transactions on Computational Biology and Bioinformatics},
  volume={19},
  number={4},
  pages={2334--2344},
  year={2021},
  publisher={IEEE}
}

@article{yang2021graphsynergy,
  title={GraphSynergy: a network-inspired deep learning model for anticancer drug combination prediction},
  author={Yang, Jiannan and Xu, Zhongzhi and Wu, William Ka Kei and Chu, Qian and Zhang, Qingpeng},
  journal={Journal of the American Medical Informatics Association},
  volume={28},
  number={11},
  pages={2336--2345},
  year={2021},
  publisher={Oxford University Press}
}

@article{zhang2023kgansynergy,
  title={KGANSynergy: knowledge graph attention network for drug synergy prediction},
  author={Zhang, Ge and Gao, Zhijie and Yan, Chaokun and Wang, Jianlin and Liang, Wenjuan and Luo, Junwei and Luo, Huimin},
  journal={Briefings in Bioinformatics},
  volume={24},
  number={3},
  pages={bbad167},
  year={2023},
  publisher={Oxford University Press}
}

@article{cheng2024hansynergy,
  title={Hansynergy: Heterogeneous graph attention network for drug synergy prediction},
  author={Cheng, Ning and Wang, Li and Liu, Yiping and Song, Bosheng and Ding, Changsong},
  journal={Journal of Chemical Information and Modeling},
  volume={64},
  number={10},
  pages={4334--4347},
  year={2024},
  publisher={ACS Publications}
}

@ARTICLE{lu2025context,
  author={Lu, Yijingxiu and Piao, Yinhua and Lee, Sangseon and Kim, Sun},
  journal={IEEE Transactions on Computational Biology and Bioinformatics}, 
  title={Context-Aware Hierarchical Fusion for Drug Relational Learning}, 
  year={2026},
  volume={23},
  number={3},
  pages={907-918},
  }

@article{zeng2025graphfusion,
    title = {GraphFusion: Integrative prediction of drug synergy using multi-scale graph representations and cell line contexts},
    journal = {Journal of Biomedical Informatics},
    volume = {171},
    pages = {104921},
    year = {2025},
    author = {Biyang Zeng and Shikui Tu and Lei Xu},
}

@ARTICLE{zhang2025mosaic,
  author={Zhang, Licai and Kang, Xiao and Yang, Xinxing and Wang, Lin and Yang, Genke and Chu, Jian},
  journal={IEEE Journal of Biomedical and Health Informatics}, 
  title={MOSAIC: A Multi-Granularity Cross-Modal Framework for Predicting Synergistic Drug Combinations in Personalized Healthcare}, 
  year={2025},
  pages={1-12},
  doi={10.1109/JBHI.2025.3605313}}

@ARTICLE{Wang2025Fusing,
  author={Wang, Xiaowen and Zhu, Hongming and Liu, Qi and Liu, Qin},
  journal={IEEE Journal of Biomedical and Health Informatics}, 
  title={Fusing Micro- and Macro-Scale Information to Predict Anticancer Synergistic Drug Combinations}, 
  year={2025},
  volume={29},
  number={3},
  pages={2297-2309},
  doi={10.1109/JBHI.2024.3500789}}

@article{liu2025mgtnsyn,
    title = {MGTNSyn: Molecular structure-aware graph transformer network with relational attention for drug synergy prediction},
    journal = {Expert Systems with Applications},
    volume = {282},
    pages = {127699},
    year = {2025},
    issn = {0957-4174},
    doi = {https://doi.org/10.1016/j.eswa.2025.127699},
    author = {Yunjiong Liu and Peiliang Zhang and Dongyang Li and Chao Che and Bo Jin},
}

@article{rafiei2023deeptrasynergy,
  title={DeepTraSynergy: drug combinations using multimodal deep learning with transformers},
  author={Rafiei, Fatemeh and Zeraati, Hojjat and Abbasi, Karim and Ghasemi, Jahan B and Parsaeian, Mahboubeh and Masoudi-Nejad, Ali},
  journal={Bioinformatics},
  volume={39},
  number={8},
  pages={btad438},
  year={2023},
  publisher={Oxford University Press}
}

@article{ying2021transformers,
  title={Do transformers really perform badly for graph representation?},
  author={Ying, Chengxuan and Cai, Tianle and Luo, Shengjie and Zheng, Shuxin and Ke, Guolin and He, Di and Shen, Yanming and Liu, Tie-Yan},
  journal={Advances in Neural Information Processing systems},
  volume={34},
  pages={28877--28888},
  year={2021}
}

@article{alam2024unlocking,
  title={Unlocking the therapeutic potential of drug combinations through synergy prediction using graph transformer networks},
  author={Alam, Waleed and Tayara, Hilal and Chong, Kil To},
  journal={Computers in Biology and Medicine},
  volume={170},
  pages={108007},
  year={2024},
  publisher={Elsevier}
}

@ARTICLE{sun2025enhancing,
  author={Sun, Bin and Du, Haoze and Hou, Shumei and Hu, Qingkai and Pang, Xiaoxiao and Wei, Dongqing and Wang, Xianfang},
  journal={IEEE Journal of Biomedical and Health Informatics}, 
  title={Enhancing Drug Synergy Combination: Integrating Graph Transformers and BiLSTM for Accurate Drug Synergy Prediction}, 
  year={2025},
  volume={29},
  number={11},
  pages={7922-7931},
  doi={10.1109/JBHI.2025.3561887}}

@article{liu2023recent,
  title={The recent progress of deep-learning-based in silico prediction of drug combination},
  author={Liu, Haoyang and Fan, Zhiguang and Lin, Jie and Yang, Yuedong and Ran, Ting and Chen, Hongming},
  journal={Drug Discovery Today},
  volume={28},
  number={7},
  pages={103625},
  year={2023},
  publisher={Elsevier}
}

@article{zhang2023smilesynergy,
  title={SMILESynergy: Anticancer drug synergy prediction based on Transformer pre-trained model},
  author={Zhang, Liqiang and Yufang, QIN and Ming, CHEN},
  journal={Journal of Biomedical Engineering},
  volume={40},
  number={3},
  pages={544},
  year={2023}
}

@article{wei2022deeplpi,
  title={DeepLPI: a novel deep learning-based model for protein--ligand interaction prediction for drug repurposing},
  author={Wei, Bomin and Zhang, Yue and Gong, Xiang},
  journal={Scientific reports},
  volume={12},
  number={1},
  pages={18200},
  year={2022},
  publisher={Nature Publishing Group UK London}
}

@article{o2016unbiased,
  title={An unbiased oncology compound screen to identify novel combination strategies},
  author={O'Neil, Jennifer and Benita, Yair and Feldman, Igor and Chenard, Melissa and Roberts, Brian and Liu, Yaping and Li, Jing and Kral, Astrid and Lejnine, Serguei and Loboda, Andrey and others},
  journal={Molecular Cancer therapeutics},
  volume={15},
  number={6},
  pages={1155--1162},
  year={2016},
  publisher={American Association for Cancer Research}
}

@article{liu2020drugcombdb,
  title={DrugCombDB: a comprehensive database of drug combinations toward the discovery of combinatorial therapy},
  author={Liu, Hui and Zhang, Wenhao and Zou, Bo and Wang, Jinxian and Deng, Yuanyuan and Deng, Lei},
  journal={Nucleic Acids Research},
  volume={48},
  number={D1},
  pages={D871--D881},
  year={2020},
  publisher={Oxford University Press}
}

@article{cheng2019network,
  title={Network-based prediction of drug combinations},
  author={Cheng, Feixiong and Kov{\'a}cs, Istv{\'a}n A and Barab{\'a}si, Albert-L{\'a}szl{\'o}},
  journal={Nature Communications},
  volume={10},
  number={1},
  pages={1197},
  year={2019},
  publisher={Nature Publishing Group UK London}
}

@article{barretina2012cancer,
  title={The Cancer Cell Line Encyclopedia enables predictive modelling of anticancer drug sensitivity},
  author={Barretina, Jordi and Caponigro, Giordano and Stransky, Nicolas and Venkatesan, Kavitha and Margolin, Adam A and Kim, Sungjoon and Wilson, Christopher J and Leh{\'a}r, Joseph and Kryukov, Gregory V and Sonkin, Dmitriy and others},
  journal={Nature},
  volume={483},
  number={7391},
  pages={603--607},
  year={2012},
  publisher={Nature Publishing Group UK London}
}

@article{landrum2013rdkit,
  title={RDKit: A software suite for cheminformatics, computational chemistry, and predictive modeling},
  author={Landrum, Greg},
  journal={Greg Landrum},
  volume={8},
  number={31.10},
  pages={5281},
  year={2013}
}

@inproceedings{su2024dual, 
    title={Dual-Channel Learning Framework for Drug-Drug Interaction Prediction via Relation-Aware Heterogeneous Graph Transformer}, 
    author={Su, Xiaorui and Hu, Pengwei and You, Zhu-Hong and Yu, Philip S. and Hu, Lun}, 
    booktitle={Proceedings of the AAAI conference on artificial intelligence}, 
    volume={38}, 
    number={1}, 
    year={2024}, 
    pages={249-256} 
}

@inproceedings{grover2016node2vec,
    author = {Grover, Aditya and Leskovec, Jure},
    title = {node2vec: Scalable Feature Learning for Networks},
    booktitle={Proceedings of the 22nd ACM SIGKDD International Conference on Knowledge Discovery and Data Mining},
    year = {2016},
    isbn = {9781450342322},
    doi = {10.1145/2939672.2939754},
    pages = {855–864},
    numpages = {10},
    location = {San Francisco, California, USA},
}

@inproceedings{xiong2020layer,
    author = {Xiong, Ruibin and Yang, Yunchang and He, Di and Zheng, Kai and Zheng, Shuxin and Xing, Chen and Zhang, Huishuai and Lan, Yanyan and Wang, Liwei and Liu, Tie-Yan},
    title = {On layer normalization in the transformer architecture},
    year = {2020},
    booktitle = {Proceedings of the 37th International Conference on Machine Learning},
    articleno = {975},
    numpages = {10},
    pages = {10524-10533},
}

@article{chen2025knowledge,
  title={Knowledge-aware contrastive heterogeneous molecular graph learning},
  author={Chen, Mukun and Wu, Jia and Pan, Shirui and Lin, Fu and Du, Bo and Gong, Xiuwen and Hu, Wenbin},
  journal={PLOS Computational Biology},
  volume={21},
  number={5},
  pages={e1013008},
  year={2025},
  publisher={Public Library of Science San Francisco, CA USA}
}

@article{prota2023structural,
  title={Structural insight into the stabilization of microtubules by taxanes},
  author={Prota, Andrea E and Lucena-Agell, Daniel and Ma, Yuntao and Estevez-Gallego, Juan and Li, Shuo and Bargsten, Katja and Josa-Prado, Fernando and Altmann, Karl-Heinz and Gaillard, Natacha and Kamimura, Shinji and others},
  journal={Elife},
  volume={12},
  pages={e84791},
  year={2023},
  publisher={eLife Sciences Publications Limited}
}

@article{coderch2012tubulin,
  title={Tubulin-based structure-affinity relationships for antimitotic Vinca alkaloids},
  author={Coderch, Claire and Morreale, Antonio and Gago, Federico},
  journal={Anti-Cancer Agents in Medicinal Chemistry},
  volume={12},
  number={3},
  pages={219-225},
  year={2012},
  publisher={Bentham Science Publishers}
}

@article{10.1371/journal.pone.0045465,
    doi = {10.1371/journal.pone.0045465},
    author = {Li, Hui AND Duan, Zhao-Wen AND Xie, Ping AND Liu, Yu-Ru AND Wang, Wei-Chi AND Dou, Shuo-Xing AND Wang, Peng-Ye},
    journal = {PLOS ONE},
    publisher = {Public Library of Science},
    title = {Effects of Paclitaxel on EGFR Endocytic Trafficking Revealed Using Quantum Dot Tracking in Single Cells},
    year = {2012},
    month = {09},
    volume = {7},
    pages = {1-13},

}

@article{deng2020role,
  title={The role of ubiquitination in tumorigenesis and targeted drug discovery},
  author={Deng, Lu and Meng, Tong and Chen, Lei and Wei, Wenyi and Wang, Ping},
  journal={Signal Transduction and Targeted Therapy},
  volume={5},
  number={1},
  pages={11},
  year={2020},
  publisher={Nature Publishing Group UK London}
}

@article{duan2021ubiquitin,
  title={Ubiquitin ligases in cancer: functions and clinical potentials},
  author={Duan, Shanshan and Pagano, Michele},
  journal={Cell Chemical Biology},
  volume={28},
  number={7},
  pages={918--933},
  year={2021},
  publisher={Elsevier}
}

@inproceedings{zhang2022predictingPLB,
  author={Zhang, Yang and Zhou, Gengmo and Wei, Zhewei and Xu, Hongteng},
  booktitle={2022 IEEE International Conference on Data Mining (ICDM)}, 
  title={Predicting Protein-Ligand Binding Affinity via Joint Global-Local Interaction Modeling}, 
  year={2022},
  volume={},
  number={},
  pages={1323-1328},
  doi={10.1109/ICDM54844.2022.00175}}

@inproceedings{Li2023discovering,
  author={Li, Xiangyu and Ovanessians, Armand and Wang, Hua},
  booktitle={2023 IEEE International Conference on Data Mining (ICDM)}, 
  title={Discovering Protein Interactions and Repurposing Drugs in SARS-CoV-2 (COVID-19) via Learning on Robust Multipartite Graphs}, 
  year={2023},
  volume={},
  number={},
  pages={289-298},
  doi={10.1109/ICDM58522.2023.00038}}



\end{document}